\documentclass[journal]{IEEEtran}

\ifCLASSINFOpdf
\else
\fi

\usepackage{cite}
\usepackage{times}
\usepackage{algorithmic}
\usepackage{array}
\usepackage{amssymb}
\usepackage{mdwmath}
\usepackage{mdwtab}
\usepackage{eqparbox}
\usepackage{cite}
\usepackage{url}
\usepackage{multicol,multirow}
\usepackage[cmex10]{amsmath}
\usepackage{makeidx}
\usepackage{diagbox}
\usepackage{rotating,soul}
\usepackage{wrapfig}
\usepackage{booktabs}
\usepackage[usenames,dvipsnames]{color}
\usepackage{ragged2e}
\usepackage{pifont}
\usepackage{overpic}
\usepackage{bm}
\makeindex

\usepackage{geometry}
\RequirePackage{silence}
\DeclareGraphicsExtensions{.pdf,.jpg,.png}

\usepackage[colorlinks,bookmarksnumbered,bookmarksopen,linkcolor=black,citecolor=blue,urlcolor=blue]{hyperref}

\def\ie{\emph{i.e.}}
\def\eg{\emph{e.g.}}
\def\etal{{\em et al.}}

\newcommand{\addFig}[1]{}
\newcommand{\addFigs}[1]{}

\graphicspath{{./figs/}}

\newlength\savedwidth

\begin{document}

%%%%%%%%% TITLE
\title{Uncertainty-aware Cross-training for Semi-supervised Medical Image Segmentation}

\author{Kaiwen Huang, Tao Zhou, Huazhu Fu, Yizhe Zhang, Yi Zhou, Xiao-Jun Wu
 	\\	 	
 	
\IEEEcompsocitemizethanks{

\IEEEcompsocthanksitem K. Huang, T. Zhou and Y. Zhang are with the School of Computer Science and Engineering, Nanjing University of Science and Technology, Nanjing 210094, China. H. Fu is with the Institute of High Performance Computing, A*STAR, Singapore. Y. Zhou is with the School of Computer Science and Engineering, Southeast University, Nanjing 214135, China. X.-J. Wu is with the School of Artificial Intelligence and Computer Science, Jiangnan University, Wuxi 214122, China. 
\IEEEcompsocthanksitem Corresponding author: Tao Zhou (taozhou.ai@gmail.com).
}
}

\markboth{}%
{Liu \MakeLowercase{\textit{et al.}}: Dynamic Feature Integration for Simultaneous Detection of Salient Object, Edge and Skeleton}

\maketitle

% \IEEEcompsoctitleabstractindextext{%

\begin{abstract}

Semi-supervised learning has gained considerable popularity in medical image segmentation tasks due to its capability to reduce reliance on expert-examined annotations. Several mean-teacher (MT) based semi-supervised methods utilize consistency regularization to effectively leverage valuable information from unlabeled data. However, these methods often heavily rely on the student model and overlook the potential impact of cognitive biases within the model. Furthermore, some methods employ co-training using pseudo-labels derived from different inputs, yet generating high-confidence pseudo-labels from perturbed inputs during training remains a significant challenge. In this paper, we propose an \textbf{U}ncertainty-aware \textbf{C}ross-training framework for semi-supervised medical image \textbf{Seg}mentation (\textbf{UC-Seg}). Our UC-Seg framework incorporates two distinct subnets to effectively explore and leverage the correlation between them, thereby mitigating cognitive biases within the model. Specifically, we present a Cross-subnet Consistency Preservation (CCP) strategy to enhance feature representation capability and ensure feature consistency across the two subnets. This strategy enables each subnet to correct its own biases and learn shared semantics from both labeled and unlabeled data. Additionally, we propose an Uncertainty-aware Pseudo-label Generation (UPG) component that leverages segmentation results and corresponding uncertainty maps from both subnets to generate high-confidence pseudo-labels. We extensively evaluate the proposed UC-Seg on various medical image segmentation tasks involving different modality images, such as MRI, CT, ultrasound, colonoscopy, and so on. The results demonstrate that our method achieves superior segmentation accuracy and generalization performance compared to other state-of-the-art semi-supervised methods. Our code and segmentation maps will be released at \href{https://github.com/taozh2017/UCSeg}{https://github.com/taozh2017/UCSeg}.

\end{abstract}

\begin{IEEEkeywords}
Medical image segmentation, semi-supervised learning, uncertainty, cross-training

\end{IEEEkeywords}

\IEEEpeerreviewmaketitle

\section{Introduction}

\IEEEPARstart{A}{ccurate} segmentation of tumors and lesions from medical images provides valuable insights to clinicians for appropriate diagnosis, disease progression assessment, and effective treatment planning~\cite{azad2024medical}. With the recent advancements in neural networks, supervised deep learning approaches have achieved remarkable performance in a diversity of medical image segmentation tasks. However, the significant progress achieved in this field can largely be attributed to the availability of large annotated datasets. Nonetheless, it is important to note that obtaining pixel-level annotation on a large scale is a time-consuming process that demands the expertise of medical professionals. To address these challenges and promote practicality, semi-supervised segmentation~\cite{huang2025learnable,chen2022semi,xie2021intra} has gained increasing attention in recent years and has become widely adopted in the field of medical image analysis. %This approach allows for leveraging both labeled and unlabeled data to train segmentation models, reducing the reliance on extensive manual annotations. 
The utilization of unlabeled data, combined with limited labeled data, enables the models to learn from additional information, improving segmentation accuracy and reducing the reliance on extensive manual annotations.

\begin{figure}[!t]
\centering
\includegraphics[width=0.9\linewidth]{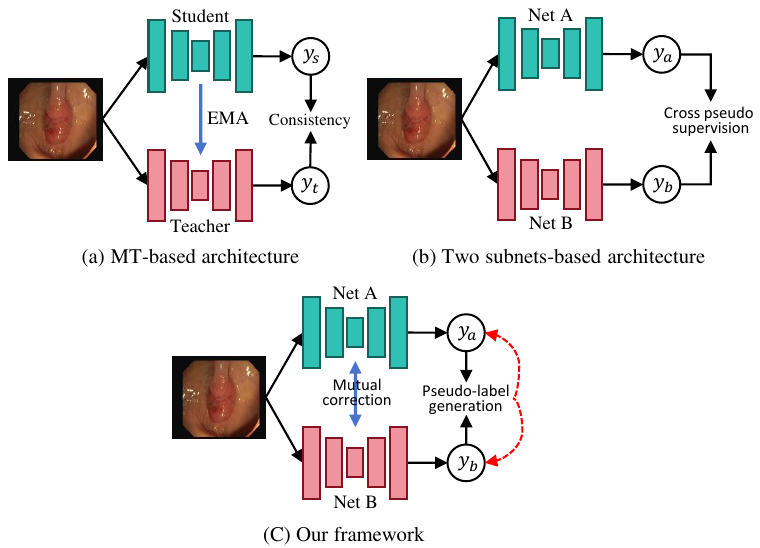}\vspace{-0.15cm}
\caption{\textbf{Motivation}: Comparison of different semi-supervised segmentation frameworks: (a) MT-based architecture~\cite{yu2019uncertainty,wang2021tripled}, (b) Two subnets-based architecture~\cite{chen2021semi}, and (c) Our framework.}
\label{fig:fig1}\vspace{-0.45cm}
\end{figure}

Several efforts have been made in semi-supervised medical image segmentation. Among existing methods, consistency learning has emerged as a powerful technique that effectively utilizes unlabeled data through unsupervised perturbation-based consistency enforcement. One notable framework in consistency learning is Mean Teacher (MT)~\cite{tarvainen2017mean}, consisting of a student network that updates parameters using gradient propagation and a teacher network that updates parameters using exponential moving average (EMA). As shown in Fig. \ref{fig:fig1}(a), MT-based methods perform label propagation by forcing consistent predictions for the perturbed inputs. Subsequently, several MT-based methods have been developed in different ways~\cite{yu2019uncertainty,shi2021inconsistency,wang2021tripled,xu2023ambiguity}. 
For example, UA-MT~\cite{yu2019uncertainty} used uncertainty information to guide the student network to learn from the meaningful and reliable targets of the teacher network gradually. CoraNet~\cite{shi2021inconsistency} proposed a model that can produce certain and uncertain regions, and the student network treats regions indicated from the teacher network with different weights. However, the limitation of the MT-based methods lies in the fact that the teacher model is a weighted mixture of the historical states of the student network, so it is heavily constrained by the student model. This constraint can lead to cognitive bias, where the model becomes biased toward certain features or regions during training, resulting in imbalanced or inaccurate predictions. Such cognitive bias can negatively affect segmentation performance, especially when the model is trained with limited labeled data. Furthermore, as discussed in \cite{wang2023mcf}, the MT architecture may exacerbate this issue by reinforcing the model's internal biases.

To address this limitation, as shown in Fig. \ref{fig:fig1}(b), the strategy of employing dual-stream networks for mutual supervision has gained popularity. This approach utilizes two distinct subnets to ensure consistency in predictions generated from the same input samples. For instance, CPS~\cite{chen2021semi} promotes similarity in model outputs for identical samples through a consistency loss. MCF~\cite{wang2023mcf} introduces two different subnets to explore and utilize the discrepancies between their outputs to correct the model's cognitive bias. Co-BioNet~\cite{peiris2023uncertainty} presents a dual-view framework based on adversarial learning, enabling one perspective to learn from the high-confidence predictions of the other by incorporating the notion of uncertainty. Additionally, some researchers have incorporated contrastive learning~\cite{basak2023pseudo, wang2023hunting} to enhance the performance of semi-supervised networks. However, these existing methods often neglect the importance of inter-subnet and intra-subnet interactions in the feature learning process. Inter-subnet interaction concerns the alignment of outputs from different subnets, while intra-subnet interaction focuses on the alignment of features within the same subnet. Overlooking these interactions can result in misaligned feature representations, which in turn restricts the model's ability to generalize and effectively utilize both labeled and unlabeled data.

This study focuses on addressing two critical challenges in semi-supervised learning: 1) facilitating effective mutual feature learning between two subnets, and 2) improving the reliability of generated pseudo-labels. To overcome these challenges, we propose a novel framework called Uncertainty-aware Cross-training for semi-supervised medical image Segmentation (UC-Seg). The UC-Seg framework incorporates two distinct structural subnets, which undergo independent parameter updates. These subnets are designed to learn from each other through a robust inter-subnet interaction, resulting in improved performance. Our framework comprises two key components: Cross-subnet Consistency Preservation (CCP) and Uncertainty-aware Pseudo-label Generation (UPG). The CCP component addresses cognitive bias within each subnet and enhances the feature representation capability by employing an Intra-subnet Feature Enhancement (IFE) strategy. This strategy ensures that each subnet can effectively learn and correct its own biases. Moreover, we propose an Inter-subnet Consistency (IC) strategy to enhance feature consistency between the two subnets. This strategy encourages the learning of common semantics from labeled and unlabeled data, promoting cross-subnet consistency during the feature learning process. Compared to existing dual-stream networks (\eg~MLRP\cite{su2024mutual}, MCF, and Co-BioNet), our method emphasizes the integration of uncertainty maps and feature embeddings for each subnet. As shown in Fig. \ref{fig:fig1}(c), this integration enables more effective alignment between the subnet predictions. Additionally, the UPG component serves as a collaborative training strategy guided by uncertainty. It leverages the segmentation results and corresponding uncertainty maps from both subnets to generate high-confidence pseudo-labels. This approach mitigates the risk of sub-networks being trapped in their biases or relying on incorrect predictions, thereby ensuring more reliable pseudo-labels. Our method facilitates effective mutual feature learning between subnets and enhances the reliability of pseudo-labels through uncertainty-aware strategies. 
% By facilitating effective mutual feature learning between subnets and enhancing the reliability of pseudo-labels through uncertainty-aware strategies, our framework presents an innovative and robust approach to improve the performance of semi-supervised learning algorithms in medical image segmentation tasks.

In summary, the contributions can be listed as follows: 
%\iffalse
\begin{itemize}

\item We propose a novel uncertainty-aware cross-training framework for semi-supervised medical image segmentation (termed UC-Seg), which can effectively employ the model bias correction and generate reliable pseudo-labels to boost the segmentation performance. 

\item A cross-subnet consistency preservation strategy is proposed to bolster the capability of feature representations and promote consistency across subnets, effectively mitigating model bias when presented in consistency-based and MT-based semi-supervised frameworks.

\item Employing a combination of segmentation maps from both subnets and their respective uncertainty maps, we present an uncertainty-aware pseudo-label generation module that produces high-confidence pseudo-labels for unlabeled data and segmentation maps for labeled data.

\item We conduct extensive experiments on diverse medical image segmentation datasets encompassing both 2D and 3D scenarios, and the results demonstrate the effectiveness and superiority of our UC-Seg over state-of-the-art (SOTA) semi-supervised segmentation methods. 

\end{itemize}

\section{Related Work}
\label{reworks}

\begin{figure*}[!t]
\centering
\includegraphics[width=1\linewidth]{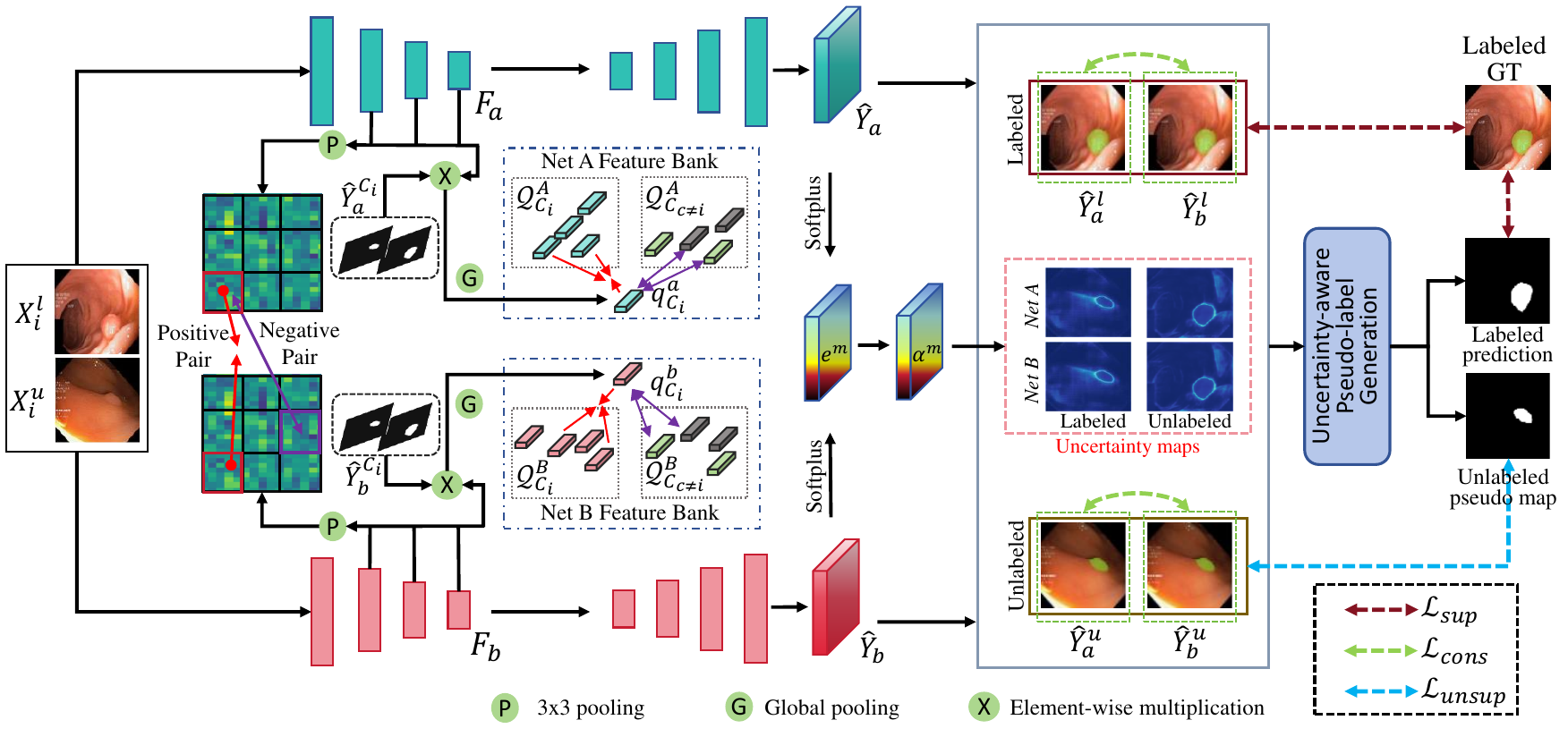}\vspace{-0.25cm}
\caption{The overall framework of the proposed UC-Seg, which consists of two key components: cross-subnet consistency preservation and uncertainty-aware pseudo-label generation. Both labeled and unlabeled data are fed into two subnets to obtain feature embeddings and predictions. Subsequently, intra-subnet feature enhancement and inter-subnet consistency strategies are implemented on the feature embeddings to enhance their discriminative ability. Additionally, the segmentation maps and their corresponding uncertainty maps from both subnets are fed into an uncertainty-aware pseudo-label generation module. This module produces high-confidence pseudo-labels for unlabeled data and refined segmentation maps for labeled data.
}\vspace{-0.15cm}
\label{fig:fig2}
\end{figure*}

\subsection{Semi-Supervised Medical Image Segmentation}

Existing semi-supervised medical image segmentation methods primarily utilize three strategies: pseudo-labeling, consistency regularization, and co-training. 

\textbf{Pseudo-labeling.} Generating pseudo-labels for unlabeled data is a well-established practice in semi-supervised learning. Therefore, it is critical to generate high-quality pseudo-labels for unlabeled data\cite{zhang2022discriminative,zeng2023ss,han2022effective,seibold2022reference}. Zhang \etal~\cite{zhang2022discriminative} employed a post-processing algorithm to refine pseudo-labels. By identifying the types and causes of segmentation errors, they proposed a dual-error correction method for unlabeled data to obtain a reliable model. 
Zeng \etal~\cite{zeng2023ss} analyzed the predictions of classification and segmentation branches to obtain reliable pseudo-labels. Additionally, pseudo-labels can be indirectly generated through label propagation, as demonstrated in~\cite{han2022effective}, which employs prototype learning to derive class representations from labeled data via mean aggregation. Subsequently, high-quality pseudo-labels are generated by computing the distances between unlabeled feature vectors and each class representation.

\textbf{Consistency Regularization.} Consistency learning aims to ensure invariance in prediction under different perturbations added to input images, which can include both input and feature perturbations. For example, Li \etal~\cite{li2020transformation} proposed to encourage network transformation consistency for unlabeled data by adding diverse noise to inputs. Shu \etal~\cite{shu2022cross} ensure consistency between the predictions of the student model with mixed inputs and the mixture of results from two teacher models. For feature perturbation, the work \cite{zheng2022double} injects random noise into the parameters of the teacher model. In addition, \cite{li2021dual} devises multiple decoders with different perturbations, ensuring consistency between the predictions of these decoders and the primary decoder.

\textbf{Co-training.} Co-training makes independent predictions for each view of the data, encouraging the model to make consistent predictions for all views~\cite{blum1998combining}. Zhao \etal~\cite{zhao2022mmgl} utilized the coronal, sagittal, and axial views of three-dimensional images as different perspectives for input. Luo \etal~\cite{luo2022semi} employed collaborative learning between two distinct network architectures, CNN and Transformer, implicitly encouraging the model to obtain consistent results from different perspectives. Peng \etal~\cite{peng2020deep} proposed to train different models using subsets of labeled data, and performed co-training on the unlabeled dataset for these models. Moreover, 
Kumari~\etal~\cite{kumari2024leveraging} proposed LLM-SegNet, a co-training framework that integrates a large language model and a unified loss function to enhance segmentation accuracy.
However, developing an effective co-training framework to produce high-quality pseudo-labels for unlabeled data is still challenging.

\subsection{Uncertainty Estimation}

Uncertainty can be categorized into two types \cite{kendall2017uncertainties}: data uncertainty and model uncertainty. Data uncertainty refers to the inherent noise present in the data, leading to inevitable errors. Additionally, model uncertainty refers to the uncertainty arising from insufficient model training, which can be mitigated through appropriate strategies. In the context of semi-supervised medical segmentation, the implementation of uncertainty-aware strategies can generate high-confidence pseudo-labels with higher confidence, which are typically more reliable~\cite{jiao2023learning, wang2022ssa}. For example, Yao \etal~\cite{yao2022enhancing} proposed a confidence-aware cross pseudo supervision algorithm to measure the variance between the original image sample and the image sample augmented by Fourier transformation, thereby improving the quality of pseudo labels. Xia \etal~\cite{xia20203d} proposed a multi-view uncertainty-aware collaborative training framework, which integrates predictions from various perspectives to generate pseudo-labels. In this study, we assess the model uncertainty of the two subnetworks and utilize the proposed uncertainty-aware pseudo-label generation strategy to mitigate the impact of uncertainty.

\section{Method}

\subsection{Overview} 

We propose a cross-training framework to fully utilize the labeled and unlabeled data to boost the semi-supervised segmentation performance. To facilitate mutual correlation learning, we propose a cross-subnet consistency preservation strategy. This strategy guides the two sub-networks to enhance the ability of feature representations and exploit the cross-subnet consistency during the feature learning process. Additionally, we present an uncertainty-aware pseudo-label generation component, which generates high-confidence pseudo-labels by incorporating uncertainty maps and prediction maps from the two subnets. The component aims to boost the accuracy of the pseudo-labels and mitigate overfitting across the entire model framework. The overall framework of our UC-Seg is illustrated in Fig.~\ref{fig:fig2}. UC-Seg consists of two sub-networks and an uncertainty-aware pseudo-label generation block. Here, we select two sub-networks, namely U-Net \cite{ronneberger2015u} and V-Net \cite{milletari2016v}, which aim to produce similar performance, denoted as subnet $\mathcal{F}_A(\cdot)$ and subnet $\mathcal{F}_B(\cdot)$. In the semi-supervised setting, the training set often consists of a small amount of labeled data, denoted as $D_L=\{(X^l_i,Y^l_i)\}^{N_l}_{i=1}$, and a large amount of unlabeled data, denoted as $D_U=\{X^u_i\}^{N_l+N_u}_{i=N_l+1}$, where $N_l\ll{N_u}$. $X_i \in \mathbb{R}^{H \times W}$ represents the input image, and $Y_i \in \{0,1\}^{C \times H \times W}$ is the ground-truth annotation with $C$ classes. The input data $X$ for each step consists of labeled data $X^l_i$ and unlabeled data $X^u_i$, and these data are fed into two sub-networks:
\begin{equation}
\left\{
\begin{aligned}
&\hat{Y}_a,  F_a = \mathcal{F}_A(X), \\
&\hat{Y}_b, F_b = \mathcal{F}_B(X),
\end{aligned}
\right.
\end{equation}
where $\hat{Y}_a$ and  $\hat{Y}_b$ represent the predictions of the two sub-networks, respectively. In addition, $F_a\in \mathbb{R}^{D\times H \times W}$ and $F_b\in \mathbb{R}^{D\times H \times W}$ denote the embedding features of the two sub-networks, respectively, where $D$ represents the dimensionality of the features.

\subsection{Cross-subnet Consistency Preservation}\label{sec:CCP}

\begin{figure}[!t]
\centering
\includegraphics[width=0.95\linewidth]{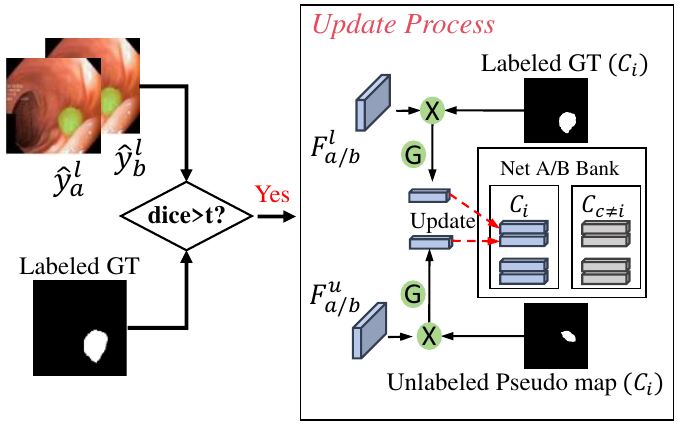}\vspace{-0.15cm}
\caption{The update process of Net A/B feature bank. The notation ``dice$>$t?" signifies whether the dice coefficient exceeds the predetermined threshold on labeled data; only when it surpasses the designated threshold will the bank be updated. Here, $C_i$ and $C_{c \neq i}$ represent the feature banks for the current class and other classes, respectively.
}
\label{fig:update_queue}\vspace{-0.35cm}
\end{figure}

In the proposed CCP component, we present an intra-subnet feature enhancement strategy, which focuses on enabling each subnet to learn discriminative feature representations for accurate segmentation of target regions. Moreover, an inter-subnet consistency strategy is proposed to enhance the feature consistency between the two subnets. By aligning their feature representations, this strategy facilitates the correction of biases within each subnet, leading to improved overall performance.

\textbf{Intra-subnet Feature Enhancement}: In medical image segmentation, a common challenge is to accurately distinguish and segment foreground regions (\eg, tumors and lesions) from complex and cluttered backgrounds. To address this, we propose leveraging contrastive learning to enhance the feature representation capability of each subnet. This strategy enables each subnet to focus more effectively on distinguishing the relevant foreground features while suppressing background noise. It also improves feature alignment within each subnet and reduces internal biases. Specifically, we construct two feature banks $\mathcal Q^{A} \in \mathbb{R}^{C \times K_f \times D}$ and $\mathcal Q^{B} \in \mathbb{R}^{C \times K_f \times D}$ to store the feature embeddings of each class for the two sub-networks, respectively, where $K_f$ denotes the maximum length of the feature bank. Subsequently, the prediction results $\hat{Y}_a \in \mathbb{R}^{C \times H \times W}$ and $\hat{Y}_b\in \mathbb{R}^{C \times H \times W}$ are scaled to align with the dimensions of their respective feature embeddings (\ie, $F_a$ and $F_b$) via bilinear interpolation. The resized predictions are then multiplied element-wise to generate the feature maps for each class. Further, the feature maps are passed through a global pooling operation to produce individual sample prototypes. To ensure that the current feature embedding is as close as possible to the distribution of samples from the same class and far from samples from different classes, we adopt the contrastive learning strategy to enhance the feature representation ability of the two subnets, which are formulated as follows:
\begin{equation}
\left\{
\begin{aligned}
&\mathcal L_q^A = -\frac{1}{C}  \sum_{i=0}^{C} \log \frac{e^{(q_{c_i}^a \cdot k_{{c_i+}}^a)/\tau}}{e^{(q_{c_i}^a \cdot k_{{c_i}+}^a)/\tau} + \sum_{j=0}^{C} e^{(q_{c_i}^a \cdot  k_{{c_j}-}^a)/\tau)}},\\
&\mathcal L_q^B = -\frac{1}{C}  \sum_{i=0}^{C} \log \frac{e^{(q_{c_i}^b \cdot k_{{c_i+}}^b)/\tau}}{e^{(q_{c_i}^b \cdot k_{{c_i}+}^b)/\tau} + \sum_{j=0}^{C} e^{(q_{c_i}^b \cdot  k_{{c_j}-}^b)/\tau)}},\\
\end{aligned}
\right.
\end{equation}
where $q_{c_i}^a \in \mathbb{R}^{D}$ and $q_{c_i}^b \in \mathbb{R}^{D}$ denote prototype representations of the predictions for the $c$-th class of individual samples, respectively. Additionally, $k_{{c_i}+}^a  \in \mathbb{R}^{D}$ and $k_{{c_i}+}^b   \in \mathbb{R}^{D}$ represent the mean values of all sample prototypes of the current class within the entire feature bank. $k_{{c_j}-}^a \in \mathbb{R}^{(C-1)\cdot k_f \times D}$ and $k_{{c_j}-}^b \in \mathbb{R}^{(C-1)\cdot k_f \times D}$ represent prototypes of samples from classes other than the current one. $\tau$ is the temperature hyperparameter. It is important to note that negative samples in this context do not necessitate averaging. The details are outlined as follows:
\begin{equation}
\left\{
\begin{aligned}
& q_{c_i}^a = \mathcal G({F}_a \odot \mathcal B(\hat{Y}_a^{c_i})),  q_{c_i}^b = \mathcal G({F}_a \odot \mathcal B(\hat{Y}_b^{c_i})),\\
& k_{{c_i}+}^a = \frac{1}{K_f} \sum_{k=0}^{K_f} \mathcal Q^{A}_{c_i},  k_{{c_i}+}^b = \frac{1}{K_f} \sum_{k=0}^{K_f} \mathcal Q^{B}_{c_i},  \\
& k_{{c_j}-}^a = \mathcal Q^A_{c \neq {c_i}}, k_{{c_j}-}^b = \mathcal Q^B_{c \neq {c_i}}, \\
\end{aligned}
\right.
\end{equation}
where $\mathcal B(\cdot)$ and $\mathcal G(\cdot)$ denote bilinear interpolation downsampling and adaptive global pooling, respectively. $\hat{Y}_a^{c_i} \in \mathbb{R}^{H \times W}$ and $\hat{Y}_b^{c_i} \in \mathbb{R}^{H \times W}$ represent the predicted results for $c$-th class in $\hat{Y}_a$ and $\hat{Y}_b$, respectively.
Similarly, $\mathcal Q^{A}_{c_i} \in \mathbb{R}^{k_f \times D}$ and $\mathcal Q^{B}_{c_i} \in \mathbb{R}^{k_f \times D}$ respectively denote the features of the $c$-th class in $\mathcal Q^{A}$ and $\mathcal Q^{B}$, while $\mathcal Q^A_{c \neq {c_i}}$ and $\mathcal Q^B_{c \neq {c_i}}$ represent the features of all samples belonging to classes other than $c$. 

As a result, the features from the same class will be closer while pulling away these features that are from different classes by such a contrastive learning strategy. Therefore, the intra-subnet feature enhancement component enables the two subnets effectively segment the tumor regions from complex backgrounds. 

\textbf{Inter-subnet Consistency}: In semi-supervised tasks, the limited availability of labeled data can lead to internal perception biases within the model during training. To mitigate the cognitive biases of each subnet, it is essential to enhance the consistency of the feature embeddings across the two subnets. This strategy reduces the likelihood of each subnet developing biases towards the input data, thereby facilitating their convergence to more accurate and unified feature representations. To achieve this, we present an inter-subnet consistency preservation strategy to constrain the soft consistency of the feature embeddings $F_a$ and $F_b$. Specifically, we employ comparative learning to implicitly enforce consistency policies. Within a batch, features embedded in patches from different networks at the same location of the same image are considered as positive sample pairs, while patches from different locations are regarded as negative samples. For 2D images, we apply a $3\times3$ pooling operation on the feature maps to partition them into patches, with each patch containing approximately one-ninth of the receptive field information from the original image. It is worth noting that for 3D data, an additional pooling operation is applied along the depth dimension.

To enhance the generalization performance across various tasks, we apply contrastive learning for consistency regularization on feature embeddings. The formula for consistency regularization of cross-sampled local feature correlations is as follows:
\begin{equation}
\mathcal L_{f} = -\frac{1}{N_p} \sum_{i=0}^{N_p} \log \frac{e^{(\mathcal P(F_a)_i \cdot \mathcal P(F_b)_i)/\tau}}{\sum_{j=0}^{N_p} e^{(\mathcal P(F_a)_i \cdot \mathcal P(F_b)_j)/\tau)}}, 
\end{equation}
where $N_p$ denotes the total number of patches, and $\mathcal P(\cdot)$ represents the pooling operation, respectively. In this case, it is important to constrain two feature maps to be consistent. This ensures that our framework reduces the discrepancy between the predictions of the two subnets and corrects cognitive biases. Note that, contrastive learning is conducted on the feature embeddings from the last three layers of the encoder.

\textbf{Embedding Sampling and Feature Bank Updating:} 

As shown in Fig.~\ref{fig:update_queue}, $\mathcal Q^{A}$ and $\mathcal Q^{B}$ are updated after each training iteration. 
Specifically, the update process of $\mathcal{Q}^{A}$ is as follows: The Dice coefficient is computed for the labeled data. If this coefficient exceeds a predefined threshold that increases linearly with the iteration count, the image features of both labeled and unlabeled data are multiplied by their corresponding ground truth and pseudo labels. These labels are resized to match the feature map size using bilinear interpolation. Subsequently, a global pooling operation is applied to obtain the individual sample prototypes, which are then stored in $\mathcal{Q}^{A}$. It should be noted that each category is calculated separately and stored in the corresponding category of $\mathcal Q^{A}$. The update process for $\mathcal{Q}^{B}$ follows the same procedure as that of $\mathcal{Q}^{A}$. When the feature bank reaches its maximum capacity, the most recently added data replaces the earliest stored data, ensuring that the feature bank maintains a fixed size.

\subsection{Uncertainty-aware Pseudo-label Generation}

Generating reliable pseudo-labels from sub-networks is a key challenge, especially when labeled data are scarce. Traditional methods often rely on preset thresholds or fixed strategies, which can lead to inaccurate results. To address this challenge, we propose a UPG strategy that yields definitive prediction maps for labeled data and accurate pseudo-labels for unlabeled data, entirely circumventing the necessity for preset thresholds. By integrating the associated uncertainty maps, the outputs from both sub-networks are effectively amalgamated. This module effectively leverages the segmentation capabilities of each sub-network, thereby enhancing overall performance and yielding significantly improved results in segmentation tasks.

\begin{figure}[!t]
\centering
\includegraphics[width=1\linewidth]{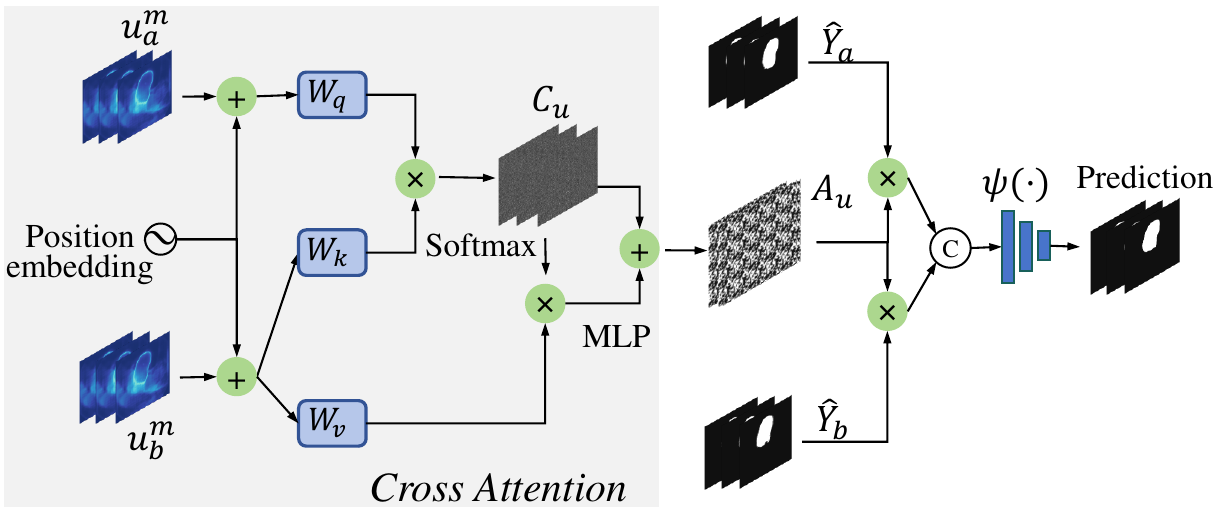}%\vspace{-0.85cm}
\caption{Architecture of the proposed Uncertainty-aware Pseudo-label Generation (UPG) strategy.
}
\label{fig:fusion_block}\vspace{-0.35cm}
\end{figure}

Fig.~\ref{fig:fusion_block} presents the details of the proposed UPG component. The core of UPG lies in computing the similarity of uncertainty maps generated by two distinct encoder-decoder pairs (\ie, $\mathcal F_A(\cdot)$ and $\mathcal F_B(\cdot)$). UPG aims to obtain more confident outputs, thereby preventing sub-networks from being adversely affected by noise and errors from predicted maps. 
Due to the high computational cost of obtaining uncertainty maps via the Monte-Carlo sampling strategy, in this study, we employ the Dirichlet distribution $Dir(\bm{\mu}^m|\bm{\alpha}^m)$ to generate uncertainty maps. This method uses subjective logic \cite{jsang2018subjective} to estimate the uncertainty of a model. It involves assigning belief mass to each class, reflecting the overall uncertainty throughout the network structure. The mass values for each class are non-negative and collectively sum up to $1$, which can be formulated by
\begin{equation}
u^m + \sum_{i=1}^{C}b_i^m = 1 \label{eq:1},
\end{equation}
where $u^m$ and $b_i^m$ denote the overall uncertainty and belief mass for the $i$-th class, respectively. Subjective logic and evidential classification \cite{bao2021evidential} associates evidence $\bm{e}^m=[e_1^m,\ldots,e_C^m]$ with the parameters of the Dirichlet distribution $\bm{\alpha}^m=[\alpha_1^m,\ldots,\alpha_C^m]$. Specifically, the relationship between $e_i^m$ and $\alpha_i^m$ is defined by $e_k^m=\alpha_k^m-1$. Therefore, $b_i^m$ and $u^m$ can be expressed as follows:
\begin{equation}
\left\{
\begin{aligned}
&b^m_i=\frac{e_i^m}{S^m} =\frac{\alpha_i^m-1}{S^m},\\
&u^m=\frac{C}{S^m} \label{eq:2},
\end{aligned}
\right.
\end{equation}
where $S^m=\sum_{i=1}^{C}(e_i^m+1)=\sum_{i=1}^{C}\alpha_i^m$ represents the Dirichlet strength. Consequently, it can be inferred that as the quantity of evidence acquired for each category grows, the assigned belief mass also increases. 

Subsequently, we employ the same strategy~\cite{zou2023evidencecap} of incorporating a \emph{Softplus} layer behind the neural network layers to ensure non-negative outputs. As a result, we generate the following evidence maps by
\begin{equation}
\left\{
\begin{aligned}
&\bm{e}_a^{m} = \ln(1 + \text{exp}{(\hat{Y}_a)}),\\
&\bm{e}_b^{m} = \ln(1 + \text{exp}(\hat{Y}_b)),
\end{aligned}
\right.
\end{equation}
where $\bm{e}_a^{m}$ and $\bm{e}_b^{m}$ denote the evidence maps for $\mathcal F_A(\cdot)$ and $\mathcal F_B(\cdot)$, respectively. In this way, a Dirichlet distribution can be obtained, which can be considered as the conjugate prior for the multinomial distribution \cite{jsang2018subjective}. Substituting $\bm{e}_a^{m}$ and $\bm{e}_b^{m}$ into Eq. (\ref{eq:2}), we can derive the uncertainty maps $u^m_a \in \mathbb{R}^{H \times W}$ and $u^m_b \in \mathbb{R}^{H \times W}$. Through the utilization of $u^m_a$ and $u^m_b  $, we construct a correlation map of uncertainty to guide the predictive weights of $\hat{Y}_a$ and $\hat{Y}_b$. Specifically, we employ a cross-attention mechanism, enabling each patch of $u^m_a$ to globally model $u^m_b$. The methodology for obtaining this instructive correlation map is as follows:
\begin{equation}
\left\{
\begin{aligned}
& C_u = \textup{softmax}\left(\frac{(W_q  u_a^m + p_e) \odot (W_k  u_b^m + p_e)^T}{\sqrt{d}}\right) \odot W_k u_b^m ,\\
& A_u =  \textup{MLP}(C_u) + C_u,\\
\end{aligned}
\right.
\end{equation}
where $A_u \in \mathbb{R}^{H \times W}$ denotes a cross-attention uncertainty map between $u^m_a$ and $u^m_b$, and $p_e$ represents the position embedding. In addition, $W_q$, $W_k$, and $W_v$ denote the learnable coefficient matrices, respectively. $d$ represents the head dimension. Finally, we obtain the output of the UPG by
\begin{equation}
\hat Y_p = \psi(Concat(A_u \odot \hat{Y}_a, A_u \odot \hat{Y}_b)),
\end{equation}
where $\psi(\cdot)$ depicts a lightweight neural network that comprises stacked Conv-BN-ReLU layers. Notably, the UPG module aims to establish correlations between the two subnets using a non-linear operation. By employing a semi-supervised learning approach, UPG identifies reliable regions within each sub-network and generates dependable outputs by adaptively integrating information based on uncertainty maps. This integration process enhances the reliability of pseudo-labels.

\subsection{Overall Loss Function}

For the supervised loss, we adopt the most widely used cross-entropy loss and dice loss in medical image segmentation, which can be defined by
\begin{equation}
\mathcal L_{ce}(\hat Y, Y) = -\frac{1}{N} \sum_{i=1}^{N} \left[ Y_i \log(\hat{Y}_i) + (1 - Y_i) \log(1 - \hat{Y}_i) \right],
\end{equation}
\begin{equation}
\mathcal L_{dice}(\hat Y, Y) = 1 - \frac{2 \sum_{i=1}^{N} Y_i \hat{Y}_i + \epsilon}{\sum_{i=1}^{N} (Y_i + \hat{Y}_i) + \epsilon},
\end{equation}
where $\hat{Y}$ and $Y$ denote the predicted maps and ground truth, respectively. $\epsilon$ is a minute value referred to as the smoothing coefficient. Following the approach of \cite{zou2023evidencecap}, we also incorporate the Calibrated Uncertainty (CU) loss into the supervision loss, which can improve the performance of the backbone network. {The loss function is defined by}
\begin{align}
    \mathcal L_{CU}(\hat Y, Y, {u}) &= -\beta_t \sum_{i,j,k \in \{\hat{Y}_{i,j,k} = Y_{i,j,k}\}} {p}_{i,j,k}log(1-{u}_{i,j,k}) \nonumber \\
& \hspace{-1cm} - (1-\beta_t) \sum_{i,j,k \in \{\hat{Y}_{i,j,k} \ne Y_{i,j,k}\}} (1-{p}_{i,j,k})log({u}_{i,j,k}),
\end{align}
where  ${p}$ and ${u}$ respectively represent the class assignment probabilities and uncertainty. $\beta_t$ denotes the annealing factor, which is defined as $\beta_t=\beta_0e^{-(ln\beta_0/T)t}$. $T$ and $t$ respectively denote the total epochs and the current epoch. In this study, $\beta_0$ is set to $0.01$. Thus, the supervised and unsupervised loss functions are denoted as $\mathcal L_s$ and $\mathcal L_u$, respectively, {which can be defined by}
\begin{align}
\mathcal L_s^A = \mathcal L_{ce}(\hat{Y}_a^l, Y^l) + \mathcal L_{dice}(\hat{Y}_a^l, Y^l) + \mathcal L_{CU}(\hat{Y}_a^l, Y^l, u_a^m),
\end{align}
\begin{equation}
\mathcal L_u^A = \mathcal L_{ce}(\hat{Y}_a^u, \hat Y_p) + \mathcal L_{dice}(\hat{Y}_a^u, \hat Y_p),
\end{equation}
where $\hat{Y}_a^l$, $\hat{Y}_a^u$ and $u_a^m$ represent the predictions on labeled data, unlabeled data, and the uncertainty map of subnet A, respectively. The distinction lies in the fact that $\mathcal L_s$ is used for labeled data, whereas $\mathcal L_u$ is employed for unlabeled data. The pseudo-labels for unlabeled data are generated by the UPG module. $\mathcal L_s^B$ and $\mathcal L_u^B$ are computed using the same method.

Moreover, the consistency constraint is also applied in the final stage of the decoder. We aim to maximize the consistency between the outputs of the two subnets when presented with the same input image. Thus, the consistency $\mathcal L_c$ is measured using CE loss by
\begin{equation}
\mathcal L_c^A = \mathcal L_{ce}(\hat{Y}_a, \hat{Y}_b),~~\mathcal L_c^B = \mathcal L_{ce}(\hat{Y}_b, \hat{Y}_a).
\end{equation}

\begin{table*}[t!]
  \centering
  \scriptsize
  \small
 \renewcommand{\arraystretch}{1.0}
   \setlength\tabcolsep{3.0pt}
  \caption{Quantitative results on the colonoscopy and ISIC-2018 datatsets.}\vspace{-0.2cm}
  \label{tab:tab1}

  \begin{tabular}{r|c|cccc|cccc|c|cccc|cccc}
  \hline

  \multicolumn{1}{c|}{\multirow{2}*{\textbf{Methods}}} &  & \multicolumn{4}{c|}{$10\%$ labeled} & \multicolumn{4}{c|}{$30\%$ labeled} &  & \multicolumn{4}{c|}{$10\%$ labeled} & \multicolumn{4}{c}{$30\%$ labeled}\\
  \cline{3-10}\cline{12-19}
   & & Dice & IoU  & 95HD & $E_{M}$ & Dice & IoU & 95HD & $E_{M}$ & & Dice & IoU  & 95HD & $E_{M}$ & Dice & IoU & 95HD & $E_{M}$\\
    \hline
  MT~\cite{tarvainen2017mean} & \multirow{15}{*}{\begin{sideways}CVC-300\end{sideways}} & 56.64 & 48.90 & 4.69 & 65.00 & 78.27 & 68.69 & 3.86 & 81.17 & \multirow{15}{*}{\begin{sideways}CVC-ClinicDB\end{sideways}}  & 78.64 & 71.02 & 3.97 & 80.00 & 83.21 & 76.92 & {3.50} & 85.25\\
  UA-MT \cite{yu2019uncertainty} & & 40.15 & 32.46 & 4.97 & 48.65 & 80.95 & 70.77 & \textcolor{blue}{3.19}  & 83.39 & & 74.51 & 67.14 & 4.05 & 76.60 & 78.89 & 72.61 & 3.68 & 81.69 \\
  DTC~\cite{luo2021semi} & & 42.81 & 33.35 & 5.42 & 50.77 & 77.44 & 67.43 & 3.32  & 80.61 & & 67.82 & 57.96 & 4.61 & 72.40 & 74.28 & 66.21 & 3.97 & 76.81\\
  CT~\cite{luo2022semi}& & 61.44 & 52.82 & 4.24 & 66.11 & 77.39 & 65.63 & 3.42 & 81.12 & & 64.70 & 55.51 & 3.93 & 68.88 & 77.54 & 70.89 & 3.85 & 80.84\\
  URPC~\cite{luo2022URPC} & & 58.83 & 50.81 & 4.30 & 64.22  & 77.05 & 66.75 & 3.65 & 80.25 & & 76.42 & 69.33 & 3.78 & 79.27 & 80.70 & 74.58 & 3.64 & 82.11\\
  MC-Net~\cite{wu2021semi} & & 69.89 & 61.58 & 3.93  & 74.47 & 81.36 & 72.11 & 3.33 & 84.49  & & 75.21 & 67.87 & 4.10 & 77.87 & 83.53 & 76.78 & 3.56 & 84.86 \\
  MC-Net+~\cite{wu2022mutual} & & 70.08 & 62.43 & 4.09 & 74.42 & 81.01 & 70.93 & 3.41 & 83.71 & & 78.23 & 70.83 & 3.94 & 81.07 & 83.77 & {77.43} & 3.58 & 85.57\\
  MCF~\cite{wang2023mcf} & & 67.58 & 57.85 & 3.91 & 73.43 & 79.51 & 69.03 & 3.27  & 83.05 & & 70.83 & 62.41 & 4.06 & 74.02 & 80.41 & 74.50 & 3.63 & 81.58\\
  SCP-Net~\cite{zhang2023self} & & 57.82 & 49.41 & 4.61 & 67.45 & 74.17 & 62.59 & 3.79 &  79.31 & & 77.41 & 69.74 & 3.97 & 80.06 & 84.71 & 77.68 & \textcolor{blue}{3.49} & 86.37\\
  CauSSL~\cite{miao2023caussl} & & 64.93 & 55.06 & 4.33 & 70.32 & 71.92 & 62.02 & 3.71 & 76.04  & & 76.23 & 68.44 & 3.99 & 78.89 & 75.18 & 68.08 & 3.89 & 77.18\\
  BS-Net~\cite{he2023bilateral} & & 65.46 & 56.31 & 4.09 & 68.83 & 80.53 & 70.99 & 3.33 & 82.46  & & 74.83 & 67.44 & 4.03 & 77.36 & 80.32 & 74.15 & 3.62 & 82.46\\
  CDMA~\cite{zhong2023semi} & & 59.44 & 48.93 & 4.34 & 66.62 & 66.38 & 52.84 & 4.77 & 73.08 & & 74.18 & 65.31 & 4.09 & 77.18 & 80.81 & 72.79 & 3.81 & 82.43\\
  UPCoL~\cite{lu2023upcol} & & 75.46 & 65.25 & 3.98 & \textcolor{blue}{81.00} & 81.36 & 72.88 & 3.46 & 83.65 & & 79.55 & 72.69 & 3.81 & 83.04 & \textcolor{blue}{85.33} & \textcolor{blue}{78.95} & 3.64 & \textcolor{blue}{87.92}\\
  BCP~\cite{bai2023bidirectional} & & \textcolor{blue}{78.33} & \textcolor{blue}{68.94} & \textcolor{blue}{3.89} & 80.52 & \textcolor{blue}{82.51} & 73.07 & 3.38 & \textcolor{blue}{84.89} & & \textcolor{blue}{82.84} & \textcolor{blue}{75.76} & \textcolor{blue}{3.64} & \textcolor{blue}{84.71} & {84.21} & 76.64 & 3.67 & 85.97\\
  CAML~\cite{gao2023correlation} & & 70.66 & 63.40 & 3.99 & 72.78 & 82.25 & \textcolor{blue}{74.14} & 3.27 & 84.03& & 77.84 & 71.71 & 3.71 & 80.87 & 80.91 & 74.93 & 3.67 & 82.87 \\
  \textbf{Ours} & & \textcolor{red}{80.63} & \textcolor{red}{70.78} & \textcolor{red}{3.60} & \textcolor{red}{83.87} & \textcolor{red}{85.07} & \textcolor{red}{76.76} & \textcolor{red}{3.14} & \textcolor{red}{86.87} & & \textcolor{red}{85.35} & \textcolor{red}{79.20} & \textcolor{red}{3.37} & \textcolor{red}{86.37} & \textcolor{red}{87.10} & \textcolor{red}{81.09} & \textcolor{red}{3.36} & \textcolor{red}{88.23}\\ 
  \hline

  MT~\cite{tarvainen2017mean} & \multirow{15}{*}{\begin{sideways}CVC-ColonDB\end{sideways}} & 47.67 & 39.30 & 5.71 & 54.62 & 63.03 & 54.63 & 4.80 & 66.24 & \multirow{15}{*}{\begin{sideways}ETIS\end{sideways}}  & 33.64 & 26.78 & 5.41 & 37.74 & 51.06 & 42.86 & 4.63 & 56.34\\
  UA-MT \cite{yu2019uncertainty} & & 41.34 & 33.49 & 5.72 & 48.63 & 59.43 & 51.30 & 4.74 & 63.25 & & 31.84 & 25.59 & 4.96 & 37.43 & 41.18 & 34.80 & 4.64 & 44.83\\
  DTC~\cite{luo2021semi} & & 35.35 & 27.17 & 6.10 & 41.82 & 54.75 & 46.18 & 5.05 & 58.69 & & 26.74 & 20.48 & 6.05 & 31.52 & 44.83 & 37.72 & 4.40 & 48.97\\
  CT~\cite{luo2022semi}& & 46.72 & 37.72 & 5.40 & 52.49 & 55.78 & 47.45 & 4.85 & 60.61 & & 28.93 & 23.15 & 4.67 & 34.00 & 41.59 & 34.40 & 4.51 & 45.93\\
  URPC~\cite{luo2022URPC} & & 46.84 & 38.98 & 5.55 & 52.95 & 60.00 & 51.89 & 4.78 & 63.22 & & 38.98 & {32.92} & {4.78} & 43.50 & 45.00 & 36.75 & 4.89 & 49.86\\
  MC-Net~\cite{wu2021semi} & & 55.69 & 46.53 & 5.24 & 61.58 & 63.65 & 55.38 & 4.71 & 66.56 & & 39.88 & 33.15 & 4.81 & 44.07 & 49.51 & 41.69 & 4.74 & 54.51\\
  MC-Net+~\cite{wu2022mutual} & & 52.36 & 44.52 & 5.47  & 58.07 & 65.94 & 57.33 & 4.62 &69.46 & & 41.71 & 33.92 & 5.20 & 48.24 & 45.07 & 38.67 & 4.69 & 48.84\\
  MCF~\cite{wang2023mcf} & & 48.72 & 40.60 & 5.25 & 53.83 & 63.24 & 55.05 & 4.64 & 67.32 & & 31.13 & 25.81 & 4.84 & 37.13 & 48.46 & 41.28& \textcolor{blue}{4.33} & 51.82\\
  SCP-Net~\cite{zhang2023self} & & 45.09 & 36.47 & 5.99 & 54.63  & 60.01 & 50.85 & 4.98 & 64.81 & & 33.73 & 27.27 & 5.07 & 38.78 & 51.24 & 41.33 & 4.85 & 57.53\\
  CauSSL~\cite{miao2023caussl} & & 48.51 & 40.89 & 5.24 & 52.53 & 57.84 & 49.36 & 4.95 & 62.44 & & 30.93 & 24.96 & 5.18 & 35.91 & 32.97 & 27.41 & 5.42 & 37.99\\
  BS-Net~\cite{he2023bilateral} & & 43.88 & 36.57 & 5.43  & 48.13 & 55.89 & 48.15 & 4.87 & 60.01 & & 31.77 & 26.26 & 5.28 & 34.57 & 38.08 & 31.65 & 4.91 & 41.88\\
  CDMA~\cite{zhong2023semi} & & 44.88 & 35.72 & 5.77 & 51.61 & 59.39 & 49.10 & 5.20 & 64.83 & & 37.23 & 28.85 & 4.87 & 41.74 & 38.29 & 29.37 & 6.07 & 48.82\\
  UPCoL~\cite{lu2023upcol} & & \textcolor{blue}{64.32} & \textcolor{blue}{54.84} & 5.17 & \textcolor{blue}{69.05} & \textcolor{blue}{69.98} & \textcolor{blue}{60.81} & 4.68 & \textcolor{blue}{73.43} & & 42.43 & 34.91 & 5.91 & 49.86 & \textcolor{red}{65.89} & \textcolor{red}{57.57} & 4.50 & \textcolor{red}{70.24}\\
  BCP~\cite{bai2023bidirectional} & & {59.91} & 51.31 & 4.93 & 64.96 & {68.11} & {58.91} & \textcolor{blue}{4.61} & 72.10 & & \textcolor{blue}{46.10} & \textcolor{blue}{37.52} & 4.95 & 51.94 & {55.28} & 46.09 & 4.52 & 60.88\\
  CAML~\cite{gao2023correlation} & & {59.86} & {52.96} & \textcolor{blue}{4.90} & 62.75 & 64.23 & 56.69 & 4.64 & 69.67 & & 33.27 & 28.94 & \textcolor{blue}{4.63} & 36.77  & 54.80 & {47.03} &  {4.36} & 59.64\\
  \textbf{Ours} & & \textcolor{red}{66.54} & \textcolor{red}{57.80} & \textcolor{red}{4.72} & \textcolor{red}{70.37} & \textcolor{red}{70.31} & \textcolor{red}{61.68} & \textcolor{red}{4.35} & \textcolor{red}{73.99} & & \textcolor{red}{51.70} & \textcolor{red}{43.87} & \textcolor{red}{4.41} & \textcolor{red}{55.41} & \textcolor{blue}{57.09} & \textcolor{blue}{48.76} & \textcolor{red}{4.25} & \textcolor{blue}{61.58}\\ 
  \hline
  
  MT~\cite{tarvainen2017mean} & \multirow{15}{*}{\begin{sideways}Kvasir\end{sideways}} & 73.99 & 64.19 & 5.37 & 79.29 & 83.74 & 75.89 & 4.53 & 86.47 & \multirow{15}{*}{\begin{sideways}ISIC-2018\end{sideways}}  & 80.86 & 72.04 & 5.32 & 84.67 & 83.70 & 75.07 & 4.96 & 86.77\\
  UA-MT \cite{yu2019uncertainty} & & 73.15 & 64.12 & 5.06  & 77.27 & 83.71 & 75.95 & 4.51 &86.55 & & 80.15 & 71.48 & 5.37 & 84.38 & 85.08 & 77.19 & 4.90 & 87.82\\
  DTC~\cite{luo2021semi} & & 66.79 & 56.00 & 5.63 & 71.71 & 77.73 & 68.78 & 4.73 & 81.26 & & 80.02 & 71.74 & 5.19 & 84.35 & 83.53 & 75.21 & 4.89 & 86.77\\
  CT~\cite{luo2022semi}& & 73.18 & 64.68 & 4.85 & 77.29 & 81.86 & 73.62 & 4.49 & 84.77 & & 81.12 & 72.43 & 5.12 & 85.08 & 83.88 & 75.06 & 4.88 & 86.92\\
  URPC~\cite{luo2022URPC} & & 79.14 & 70.53 & 4.65 & 83.21 & 84.47 & 76.63 & 4.34 & 87.66 & & 79.38 & 70.36 & 5.37 & 83.61 & 83.18 & 74.85 & 4.91 & 86.61\\
  MC-Net~\cite{wu2021semi} & & 78.03 & 69.21 & 4.89 & 82.03 & 85.51 & 78.07 & 4.46 & 88.08 & & 80.98 & 72.03 & 5.31 & 84.95 & 83.21 & 74.42 & 4.98 & 86.16\\
  MC-Net+~\cite{wu2022mutual} & & 78.57 & 69.59 & 4.84 & 83.05 & 86.18 & 79.57 & {4.21}  &89.21 & & 80.72 & 71.75 & 5.37 & 84.51 & 84.37 & 76.50 & 4.85 & 87.42\\
  MCF~\cite{wang2023mcf} & & 74.01 & 64.65 & 4.99 & 79.22 & 83.25 & 75.42 & 4.46  & 86.22 & & 80.76 & 71.88 & 5.09 & 84.44 & 83.50 & 75.13 & 4.86 & 86.62\\
  SCP-Net~\cite{zhang2023self} & & 74.42 & 64.64 & 5.17 & 78.93 & 84.29 & 76.45 & 4.67 & 86.81  & & 78.95 & 69.80 & 5.45 & 83.19 & 83.52 & 74.95 & 5.04 & 86.73\\
  CauSSL~\cite{miao2023caussl} & & 76.48 & 67.61 & 4.93 & 81.05  & 81.03 & 73.20 & 4.67 & 84.50 & & 79.73 & 71.19 & 5.34 & 84.19 & 84.53 & 75.88 & 4.93 & 87.23\\
  BS-Net~\cite{he2023bilateral} & & 77.18 & 68.22 & 4.99 & 80.66 & 82.50 & 74.98 & 4.45 & 85.52  & & 81.70 & 73.68 & 5.05 & 85.30 & 84.22 & 75.83 & 4.85 & 86.99\\
  CDMA~\cite{zhong2023semi} & & 74.98 & 65.06 & 5.07 & 78.82 & 82.43 & 73.47 & 4.63 & 85.42 & & 76.70 & 66.58 & 5.84 & 81.83 & 83.27 & 74.88 & 4.78 & 86.69\\
  UPCoL~\cite{lu2023upcol} & & \textcolor{blue}{82.12} & \textcolor{blue}{74.40} & \textcolor{blue}{4.51} & \textcolor{blue}{84.87} & 86.72 & 80.15 & \textcolor{blue}{4.14} & 88.80 & & \textcolor{blue}{84.87} & \textcolor{blue}{76.63} & 5.11 & \textcolor{blue}{87.22}  & \textcolor{blue}{86.55} & \textcolor{blue}{78.43} & \textcolor{blue}{4.74} & \textcolor{blue}{88.61}\\
  BCP~\cite{bai2023bidirectional} & & 79.64 & 70.44 & 4.85 & 82.46 & 85.99 & 78.52 & 4.56 & 88.38 & & {83.98} & {75.15} & 5.00 & 87.01 & {85.35} & {77.56} &{4.84} & 88.15\\
  CAML~\cite{gao2023correlation} & & {80.66} & {73.27} & {4.58} & 83.96 & \textcolor{blue}{87.25} & \textcolor{blue}{80.58} & 4.22 & \textcolor{blue}{89.08} & & 81.89 & 73.66 & \textcolor{blue}{4.96} & 85.36 & 83.65 & 75.57 & 4.86 & 86.62\\
  \textbf{Ours} & & \textcolor{red}{86.81} & \textcolor{red}{79.51} & \textcolor{red}{4.10} & \textcolor{red}{88.93} & \textcolor{red}{88.27} & \textcolor{red}{81.69} & \textcolor{red}{3.97} & \textcolor{red}{90.16} & & \textcolor{red}{86.37} & \textcolor{red}{78.84} & \textcolor{red}{4.62} & \textcolor{red}{88.97} & \textcolor{red}{87.40} & \textcolor{red}{79.52} & \textcolor{red}{4.52} & \textcolor{red}{89.48}\\ 
  \hline
  \end{tabular}
  %\end{adjustbox}
\end{table*}

Finally, the loss functions for the two sub-networks are as follows:
\begin{equation}
\left\{
\begin{aligned}
& \mathcal L^A =  \mathcal L_s^A + \lambda_u \mathcal L_u^A + \lambda_q \mathcal L_q^A  + \lambda_c \mathcal L_c^A,\\
& \mathcal L^B = \mathcal L_s^B  + \lambda_u \mathcal L_u^B + \lambda_q \mathcal L_q^B +  \lambda_c \mathcal L_c^B,\\
\end{aligned}
\right.
\end{equation}
where $\lambda_u$, $\lambda_q$, and $\lambda_c$ are parameters that control the weights of $\mathcal L_u$, $\mathcal L_q$, and $\mathcal L_c$ respectively. They are used to balance the proportions of weights among the various losses. In this study, $\lambda_q$ and $\lambda_c$ are set to 0.2 and 0.5, respectively. We used a Gaussian warm-up function $\lambda_u(t) = \beta * e^{-5(1-t/t_{max})^2}$ in unsupervised loss \cite{laine2016temporal}, where $\beta$ is set to 15. The UPG block is trained only on labeled data, so it only incurs a supervised loss, {as shown below:}
\begin{equation}
\mathcal L^P = \mathcal L_{dice}(\hat{Y}_p, Y^l) + \mathcal L_{ce}(\hat{Y}_p, Y^l) + \mathcal L_{CU}(\hat{Y}_p, Y^l, u_p^m),
\end{equation}
where $u_p^m$ denotes the uncertainty of UPG. It is worth noting that the gradient back-propagation of the loss function in UPG does not affect the parameters in the sub-network, only the parameters of UPG will be updated. In conclusion, the overall loss function is expressed as:
\begin{equation}
\mathcal L_{total} = (\mathcal L^A + \mathcal L^B) / 2 + \mathcal L^P + \lambda_f \mathcal L_f,
\end{equation}
where $\lambda_f$ is set to 0.1, and $\mathcal L_f$ has been elaborated upon in Section~\ref{sec:CCP}.

\begin{table*}[t!]
  \centering
  \scriptsize
  \small
   \renewcommand{\arraystretch}{1.0}
   \setlength\tabcolsep{3.0pt}
  \caption{Quantitative results on three segmentation tasks.} \vspace{-0.2cm} %with 
  \label{tab:tab2}

  \begin{tabular}{r|c|cccc|cccc|c|cccc|cccc}
  \hline

  \multicolumn{1}{c|}{\multirow{2}*{\textbf{Methods}}} &  & \multicolumn{4}{c|}{$10\%$ labeled} & \multicolumn{4}{c|}{$30\%$ labeled} &  & \multicolumn{4}{c|}{$10\%$ labeled} & \multicolumn{4}{c}{$30\%$ labeled}\\
  \cline{3-10}\cline{12-19}
   & & Dice & IoU  & 95HD & {$E_{M}$} & Dice & IoU & 95HD & {$E_{M}$} & & Dice & IoU  & 95HD & {$E_{M}$} & Dice & IoU & 95HD & {$E_{M}$}\\
    \hline
  MT~\cite{tarvainen2017mean} & \multirow{15}{*}{\begin{sideways}DDTI\end{sideways}} & 31.59 & 24.32 & 6.79 & 37.66 & 47.62 & 38.17 & 6.24 &  52.19 & \multirow{15}{*}{\begin{sideways}TN3K\end{sideways}}  & 73.09 & 62.95 & 4.96  & 77.12 & 78.88 & 69.71 & 4.53 &  81.71 \\
  UA-MT \cite{yu2019uncertainty} & & 34.96 & 27.31 & 6.73 &  40.39 & 49.76 & 39.85 & 6.07  & 54.56 & & 74.73 & 64.23 & 4.83 &  78.87 & 78.42 & 69.16 & 4.53 &  81.54\\
  DTC~\cite{luo2021semi} & & 27.13 & 19.67 & 7.35 & 35.91 & 38.53 & 28.05 & 6.87 & 48.27 & & 62.67 & 50.50 & 5.61 & 68.98 & 64.48 & 51.66 & 5.74 & 70.32\\
  {CT}~\cite{luo2022semi}& & 33.71 & 26.21 & 6.69 & 40.58 & 46.45 & 36.79 & 6.38 & 51.32 & & 70.58 & 60.04 & 5.01 & 75.33 & 76.18 & 66.62 & 4.63 & 79.50\\
  URPC~\cite{luo2022URPC} & & 31.37 & 24.23 & 6.88 &  36.29 & 56.01 & 45.26 & 5.92 &  61.21  & & 74.42 & 64.17 & 4.88 &  78.66 & 79.19 & 70.00 & 4.56 & 82.04 \\
  MC-Net~\cite{wu2021semi} & & 29.42 & 23.12 & 7.15 & 36.36 & 45.36 & 36.60 & 6.20  &  50.04 & & 71.45 & 60.72 & 5.28  & 75.99 & 78.04 & 68.72 & 4.65 &  80.85 \\
  MC-Net+~\cite{wu2022mutual} & & 41.14 & 32.34 & 6.52 & 45.95 & 47.26 & 38.61 & 6.09 & 52.50 & & 75.05 & 65.13 & 5.10 & 78.93  & 79.70 & 70.26 & 4.50  & 82.50\\
  MCF~\cite{wang2023mcf} & & \textcolor{blue}{57.69} & \textcolor{blue}{46.84} & \textcolor{blue}{5.71} & \textcolor{blue}{63.64} & 55.09 & 44.08 & 5.97 &  60.66 & & 74.50 & 64.46 & 4.79 & 78.44 & 78.04 & 68.79 & 4.43 & 80.94 \\
  SCP-Net~\cite{zhang2023self} & & 36.23 & 27.94 & 6.86 &  41.24 & 46.24 & 36.87 & 6.16 & 50.77 & & 73.26 & 63.30 & 5.03 &  77.31 & 78.36 & 69.19 & 4.57 & 81.35\\
  CauSSL~\cite{miao2023caussl} & & 41.88 & 32.76 & 6.52 & 48.64 & 46.12 & 36.51 & 6.22 &  52.78 & & 70.89 & 60.22 & 5.17 &  74.92 & 73.93 & 63.63 & 4.97 & 77.83 \\
  BS-Net~\cite{he2023bilateral} & & 31.41 & 24.82 & 6.74 &  36.59 & 40.55 & 32.42 & 6.33 &  45.16 & & 73.70 & 64.22 & 4.77 &  77.83 & 79.25 & 70.60 & 4.39 & 81.57\\
  CDMA~\cite{zhong2023semi} & & 46.42 & 36.17 & 6.28 & 53.70 & 54.68 & 42.48 & 6.34  & 61.88 & & 72.97 & 61.54 & 5.12 & 77.68 & 76.24 & 65.60 & 4.81 & 80.14\\
  UPCoL~\cite{lu2023upcol} & & 53.47 & 41.62 & 6.36 & 58.83 & \textcolor{blue}{57.47} & \textcolor{blue}{45.34} & 6.65 & \textcolor{blue}{64.25} & & 76.81 & 66.39 & 4.91 &  80.49 & \textcolor{blue}{80.71} & 71.18 & 4.55 &  \textcolor{blue}{83.73}\\
  BCP~\cite{bai2023bidirectional} & & 57.45 & 44.78 & 5.97  & 62.75 & 52.30 & 41.00 & 6.10 & 57.25 & & 75.36 & 64.65 & 5.04 & 79.50 & 76.70 & 66.59 & 4.82 &  80.00\\
  CAML~\cite{gao2023correlation} & & 42.45 & 34.50 & 6.33 & 47.01 & {55.94} & {46.28} & \textcolor{blue}{5.73} & 62.70 & & \textcolor{blue}{77.78} & \textcolor{blue}{68.80} & \textcolor{red}{4.47} & \textcolor{blue}{80.82} & {80.15} & \textcolor{blue}{71.30} & \textcolor{blue}{4.40} &  82.87\\
  \textbf{Ours} & & \textcolor{red}{58.12} & \textcolor{red}{47.50} & \textcolor{red}{5.66} & \textcolor{red}{64.74} & \textcolor{red}{62.00} & \textcolor{red}{50.96} & \textcolor{red}{5.54} & \textcolor{red}{67.67} & & \textcolor{red}{80.64} & \textcolor{red}{71.32} & \textcolor{blue}{4.53} & \textcolor{red}{83.62} & \textcolor{red}{82.17} & \textcolor{red}{73.34} & \textcolor{red}{4.33} & \textcolor{red}{84.53}\\ 
  \hline

  MT~\cite{tarvainen2017mean} & \multirow{15}{*}{\begin{sideways}Brain MRI\end{sideways}} & 63.52 & 54.44 & 3.61 & 66.89 & 63.55 & 54.74 & 3.53 & 67.76 & \multirow{15}{*}{\begin{sideways}ACDC\end{sideways}}  & 81.89 & 71.10 & 12.48 & 54.55 & 87.06 & 77.94 & 11.17 & 87.84\\
  UA-MT \cite{yu2019uncertainty} & & 61.05 & 51.19 & 3.67 & 65.30 & 68.00 & 58.99 & 3.41 &  71.07 & & 81.65 & 70.64 & 6.88 & 83.25 & 87.75 & 78.98 & 4.78 & 88.30\\
  DTC~\cite{luo2021semi} & & 60.96 & 51.88 & 3.64 & 63.70 & 63.92 & 54.95 & 3.55 & 66.36 & & 84.29 & 73.92 & 12.81 & - & 85.75 & 75.99 & 5.08 & 86.47\\
  {CT}~\cite{luo2022semi}& & 57.13 & 48.38 & 2.80 & 61.06 & 61.84 & 52.68 & 3.78 & 66.39 & & 86.40 & - & 8.60 & - & 88.14 & 79.42 & 6.26 & 88.62\\
  URPC~\cite{luo2022URPC} & & 61.87 & 52.87 & 3.64 & 64.87 & 70.74 & 59.81 & 3.27 & 75.83 & & 83.10 & 72.41 & 4.84 & - & 88.36 & 79.96 & {3.04} & 88.94\\
  MC-Net~\cite{wu2021semi} & & 64.57 & 55.30 & 3.56 & 68.50 & 73.62 & 63.17 & 3.17  & 77.66 & & 86.44 & 77.04 & 5.50 & - & 88.04 & {81.01} & 3.54 & 89.56\\
  MC-Net+~\cite{wu2022mutual} & & 65.10 & 55.86 & 3.57 & 67.65 & 66.96 & 58.10 & 3.43 & 70.79 & & 85.30 & 75.76 & 12.11 & \textcolor{blue}{86.46} & 88.39 & 80.08 & 6.38 & 89.05\\
  MCF~\cite{wang2023mcf} & & 57.64 & 48.13 & 3.70 & 61.94 & 63.64 & 54.31 & 3.49 & 68.88 & & 83.23 & 72.93 & 12.79 & 83.94  & 86.22 & 76.73 & 7.42 & 88.34 \\
  SCP-Net~\cite{zhang2023self} & & 60.65 & 51.68 & 3.68 & 64.08 & 66.23 & 56.91 & 3.52 & 69.28 & & 80.90 & 69.64 & 15.64 & 84.41 & 89.03 & 80.83 & 6.61 & 89.64 \\
  CauSSL~\cite{miao2023caussl} & & 61.73 & 52.21 & 3.64 & 65.77 & 64.65 & 55.58 & 3.48 & 68.40 & & 86.80 & 77.48 & 5.73 & - & 85.11 & 75.00 & 9.80 & 85.77\\
  BS-Net~\cite{he2023bilateral} & & 53.48 & 44.51 & 3.84 & 59.00 & 62.03 & 53.27 & 3.53 & 66.10 & & 77.51 & 65.38 & 17.49 & 79.47 &  \textcolor{blue}{89.31} & \textcolor{blue}{81.24} & \textcolor{blue}{2.10} & \textcolor{blue}{89.71}\\
  CDMA~\cite{zhong2023semi} & & 58.02 & 48.46 & 3.74 &  61.87 & 62.29 & 53.29 & 3.56  & 66.41 & & 83.77 & 73.00 & 4.94 & 84.37 & 84.68 & 74.46 & 6.27 & 85.72\\
  UPCoL~\cite{lu2023upcol} & & \textcolor{blue}{72.36} & \textcolor{blue}{61.48} & \textcolor{blue}{3.34} &  \textcolor{blue}{77.85} & 75.47 & 64.78 & 3.26 & 79.81 & & 84.48 & 74.14 & 8.09 & 85.47 & 88.19 & 79.65 & 3.17 & 88.83\\
  BCP~\cite{bai2023bidirectional} & & {69.65} & {60.10} & {3.44} & 71.47 & \textcolor{blue}{78.46} & \textcolor{blue}{68.46} & \textcolor{blue}{3.03} & \textcolor{blue}{81.13} & & \textcolor{blue}{88.84} & \textcolor{blue}{80.62} & \textcolor{blue}{3.98} & - & {88.61} & 80.12 & 4.71 & 88.85 \\
  CAML~\cite{gao2023correlation} & & 62.24 & 53.69 & 3.58 & 65.29 & 76.12 & 65.91 & 3.13 & 80.53 & &  81.90 & 70.60 & 6.46 &  83.21 & 86.85 & 77.63 & 5.08 & 87.62 \\
  \textbf{Ours} & & \textcolor{red}{74.33} & \textcolor{red}{63.54} & \textcolor{red}{3.23} & \textcolor{red}{79.67} & \textcolor{red}{79.59} & \textcolor{red}{69.07} & \textcolor{red}{3.02} & \textcolor{red}{82.63} & & \textcolor{red}{88.86} & \textcolor{red}{80.66} & \textcolor{red}{1.58} & \textcolor{red}{89.41} & \textcolor{red}{90.47} & \textcolor{red}{83.19} & \textcolor{red}{1.29} & \textcolor{red}{90.87}\\ 
  \hline
  
  \end{tabular}
\end{table*}

\section{Experiments}
\label{Experiments}

\subsection{Datasets}

\textbf{Colonoscopy Datasets}: Five colonoscopy image datasets are used in the experiment,  namely ETIS \cite{silva2014toward}, CVC-ClinicDB \cite{bernal2015wm}, CVC-ColonDB \cite{tajbakhsh2015automated}, CVC-300 \cite{vazquez2017benchmark}, and Kvasir \cite{jha2020kvasir}. Following the previous setting~\cite{zhou2023cross}, 1,450 images from the CVC-ClinicDB and Kvasir datasets are selected to form the training set, while the remaining images from the two datasets (\ie, CVC-ClinicDB and Kvasir) and three datasets (\ie, CVC-ColonDB, CVC-300, and ETIS) are for test.

\textbf{Skin Lesion Dataset}: The ISIC 2018 \cite{codella2019skin} dataset, released by the International Skin Imaging Collaboration (ISIC), comprises a substantial collection of dermoscopy images. This dataset encompasses a total of 2,594 images for training and 1,000 images for test. 

\textbf{Ultrasound Thyroid Nodule Datasets}: Three ultrasound datasets are employed in the experiments, namely TN3K \cite{gong2023thyroid}, TG3K \cite{wunderling2017comparison}, and DDTI \cite{pedraza2015open}. TN3K consists of 3,493 ultrasound images with pixel-level annotations, with 2,879 training images and 614 testing images. DDTI comprises 637 ultrasound images from a single device. The TG3K dataset originates from 16 ultrasound videos, and 3,585 ultrasound images are extracted from these videos~\cite{gong2023thyroid}. Following the experimental setting in~\cite{gong2023thyroid}, we have 6,464 images (2,879 images from TN3K and 3,585 ones from TG3K) to form the training and validation set, while 614 images from TN3K and 637 images from DDTI are taken as testing sets.

\textbf{Brain Tumor Dataset}: This dataset encompasses magnetic resonance (MR) images of the brain, accompanied by manually generated segmentation masks highlighting FLAIR abnormalities. Originating from The Cancer Imaging Archive (TCIA), these images are linked to 110 patients featured in the Cancer Genome Atlas (TCGA) collection of lower-grade gliomas. The dataset comprises 1,373 images, where 1,036, 54, and 283 images are used for training, validation, and testing, respectively. 

\begin{figure*}[!t]
\centering
\includegraphics[width=0.96\linewidth]{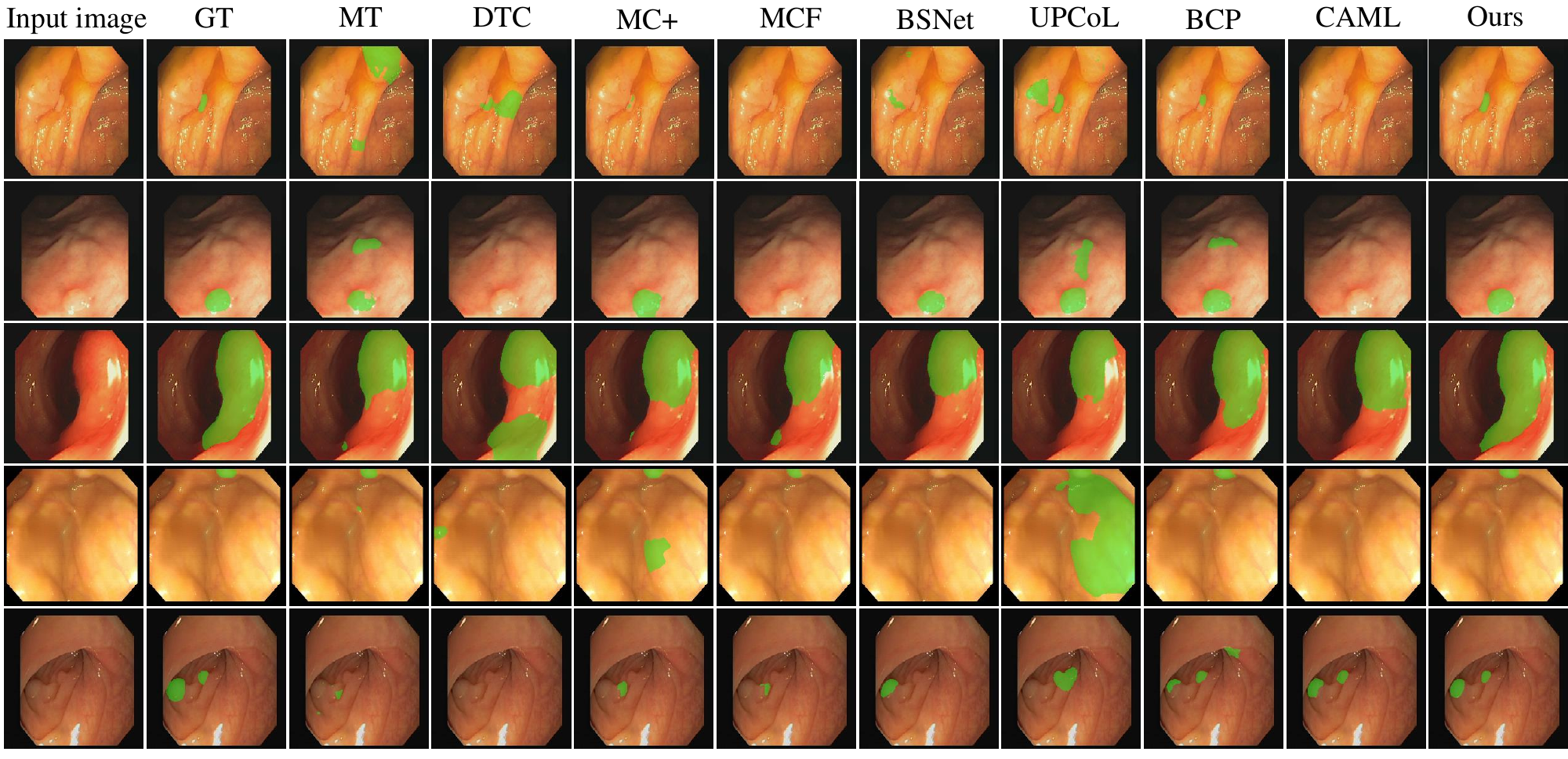}\vspace{-0.2cm}
\caption{Qualitative visualization of polyp segmentation results comparing our model with other representative methods with $30\%$ labeled data, encompassing samples of varying polyp sizes. 
}\vspace{-0.2cm}
\label{fig:tumor_result}
\end{figure*}

\begin{table*}[t!]
  \centering
  \scriptsize
  \small
   \renewcommand{\arraystretch}{1.0}
   \setlength\tabcolsep{3.2pt}
  \caption{Quantitative results on the LA and NIH Pancreas datasets.} \vspace{-0.2cm} 
  \label{tab:3D_Res}
  \begin{tabular}{r|c|cccc|cccc|c|cccc|cccc}
  \hline
  %\rowcolor{mygray}
  \multicolumn{1}{c|}{\multirow{2}*{\textbf{Methods}}} &  & \multicolumn{4}{c|}{$5\%$ / 4 labeled data} & \multicolumn{4}{c|}{$10\%$ / 8 labeled data} &  & \multicolumn{4}{c|}{$10\%$ / 6 labeled data} & \multicolumn{4}{c}{$20\%$ / 12 labeled data}\\
  \cline{3-10}\cline{12-19}
   & & Dice & IoU  & 95HD & ASD & Dice & IoU & 95HD & ASD & & Dice & IoU  & 95HD & ASD & Dice & IoU & 95HD & ASD\\
    \hline
  UAMT~\cite{tarvainen2017mean} & \multirow{11}{*}{\begin{sideways}Left Atrium\end{sideways}} & 82.26 & 70.98 & 13.71 & 3.82   & 87.79 & 78.39 & 8.68 & 2.12 & \multirow{11}{*}{\begin{sideways}Pancreas\end{sideways}}  & 68.70 & 54.65 & 13.89 & 3.23  & 77.26 & 63.82 & 11.90 & 3.06 \\
  DTC~\cite{luo2021semi} & &   81.25 & 69.33 & 14.90 & 3.99   & 87.51 & 78.17 & 8.23 & 2.36 & & 66.27 & 52.07 & 15.00 & 4.44   & 76.75 & 63.78 & 8.52 & 1.72\\
  URPC~\cite{luo2022URPC} & &   82.48 & 71.35 & 14.65 & 3.65 & 86.92 & 77.03 & 11.13 & 2.28  & & 64.73 & 49.62 & 21.90 & 7.73         & 79.09 & 65.99 & 11.68 & 3.31 \\
  SASSNet~\cite{li2020shape} & &  81.60 & 69.63 & 16.16 & 3.58        & 87.54 & 78.05 & 9.84 & 2.59 & & 66.52 & 52.23 & 17.11 & 2.25         & 77.11 & 64.24 & 11.41 & 2.79 \\
  MC-Net~\cite{wu2021semi} &  & 83.59 & 72.36 & 14.07 & 2.70   & 87.62 & 78.25 & 10.03 & 1.82 & & 69.07 & 54.36 & 14.53 & 2.28   & 78.27 & 64.75 & 8.36 & 2.25\\
  MC-Net+~\cite{wu2022mutual} & &   81.60 & 69.42 & 14.32 & 3.16    & 88.89 & 80.15 & 8.01 & 1.90 & & 70.00 & 55.66 & 16.03 & 3.87         & 79.37 & 66.83 & 8.93 & 1.41\\
  MCF~\cite{wang2023mcf} & & 79.02 & 66.90 & 16.34 & 4.46 & 87.06 & 77.83 & 7.81 & 2.67 & & 69.00 & 54.80 & 14.23 & 1.90 & 75.00 & 61.27 & 11.59 & 3.27  \\
  Co-BioNet \cite{peiris2023uncertainty} & &  84.38 & 74.67 & 8.33 & 2.38         & 89.20 & 80.68 & \textcolor{red}{6.44} & 1.90 & &  77.89 & 64.79 & \textcolor{blue}{8.81} & \textcolor{red}{1.39}         & 82.22 & 70.24 & 7.71 & {1.16} \\
  CauSSL~\cite{miao2023caussl} & &  78.82 & 65.99 & 19.76 & 5.84   & 89.05 & 80.40 & 7.60 & 1.95 & &  72.89 & 58.06 & 14.19 & 4.37    & 80.92 & 68.26 & 8.11 & 1.53\\
  BS-Net~\cite{he2023bilateral} & &  83.28 & 72.01 & 14.28 & 3.56      & 86.18 & 76.33 & 11.47 & 3.02 & &  64.61        &  50.02  &  24.74  &  5.28 & 78.93   & 65.75       &8.49  &  2.20\\
  UPCoL~\cite{lu2023upcol} & & 82.08  & 70.57 & 15.80 & 3.62 & 88.51 & 79.66 & 8.09 & 2.03 & & 67.92 & 54.02 & 20.33 & 4.97 & 81.78 & 69.66 & 3.78 & \textcolor{red}{0.63}  \\
  BCP~\cite{bai2023bidirectional} & &  \textcolor{blue}{88.02} & \textcolor{blue}{78.72} & \textcolor{blue}{7.90} & \textcolor{blue}{2.15}              & \textcolor{blue}{89.62} & \textcolor{blue}{81.31} & 6.81 & \textcolor{blue}{1.76} & & \textcolor{blue}{79.19} & \textcolor{blue}{66.12} & 9.31 & 1.53     & \textcolor{blue}{82.91} & \textcolor{blue}{70.97} & \textcolor{blue}{6.43} & 2.25 \\
  CAML~\cite{gao2023correlation} & & 87.34 & 77.65 & 9.76 & 2.49 & 89.62 & 81.28 & 8.76 & 2.02 & & 71.21 & 56.32 & 16.89& 5.92  & 79.81 & 67.35 & 8.22 & 2.27  \\
  \textbf{Ours} & & \textcolor{red}{90.36} & \textcolor{red}{82.48} &  \textcolor{red}{6.35} & \textcolor{red}{1.76}                  & \textcolor{red}{90.90} & \textcolor{red}{83.41} & \textcolor{blue}{6.50} & \textcolor{red}{1.62} & & \textcolor{red}{81.71} & \textcolor{red}{69.50} & \textcolor{red}{4.99} & \textcolor{blue}{1.44}         & \textcolor{red}{83.84} & \textcolor{red}{72.50} & \textcolor{red}{3.79} & \textcolor{blue}{1.07}\\ 
  \hline
  \end{tabular}
  %\end{adjustbox}
\end{table*}

\textbf{Automatic Cardiac Diagnosis Challenge:} ACDC~\cite{bernard2018deep} is a challenge held during MICCAI 2017. The dataset comprises four classes: background, right ventricle, left ventricle, and myocardium. There are a total of 100 patients' scans in the dataset. Following the allocation method proposed by~\cite{bai2023bidirectional}, the dataset is divided into training, validation, and test sets, comprising 70, 10, and 20 scans, respectively.

\textbf{Left Atrium (LA) Dataset}: The LA dataset \cite{xiong2021global} serves as the benchmark dataset for the 2018 atrial segmentation challenge. It consists of 100 3D gadolinium-enhanced magnetic resonance image volumes with labels.  We still strictly adhere to the same configuration as in previous studies, employing a data-splitting strategy with 80 samples for training and 20 samples for testing in our experiments.

\textbf{NIH Pancreas CT Dataset}: The Dataset \cite{roth2015deeporgan} comprises 82 abdominal enhanced 3D CT scans obtained from a cohort of 53 males and 27 females, with participants' ages ranging from 18 to 76 years. We adopted the identical data segmentation approach as Co-bioNet \cite{peiris2023uncertainty}, utilizing 62 samples for training and allocating 20 samples for testing purposes.

During the training process, for the division of labeled data, we randomly selected $10\%$ and $30\%$ of the data for experiments in the colonoscopy, skin lesion, ultrasound thyroid nodule, and brain tumor segmentation tasks. For the ACDC, Left Atrium (LA), and NIH Pancreas CT datasets, we follow the data allocation strategies from previous works~\cite{bai2023bidirectional} to ensure a fair experimental comparison.

\subsection{Implementation Details}

All experiments were conducted in a PyTorch 1.8.1 environment with CUDA 11.2 and an NVIDIA 3090 GPU. To facilitate comparisons with existing models, U-Net \cite{ronneberger2015u} and V-Net \cite{milletari2016v} are adopted as the backbone networks for the two sub-nets in our model. Throughout the entire training process, the batch size was set to 24, consisting of 12 labeled and 12 unlabeled images. We employed the SGD optimizer with a learning rate of 0.01, momentum of 0.9, and weight decay of 1e-4 for 50,000 iterations. The maximum length of feature bank $\mathcal Q^{A}$ and $\mathcal Q^{B}$ is set to 1440 tokens. 
During the feature bank update, the value of $t$ changes with iterations. For 2D data, it is set to $0.4 + 0.56(t/t_{max})$, and for 3D data, it is set to $0.2 + 0.56(t/t_{max})$. 
Additionally, we adopt a widely used copy-paste \cite{bai2023bidirectional, yun2019cutmix, zhang2017mixup} data augmentation strategy, alleviating the distributional differences between labeled and unlabeled data during the training process. 

\subsection{Comparison With State-of-the-art Methods}

We compare the proposed method with current top-performing approaches in the field of semi-supervised medical image segmentation, including MT~\cite{tarvainen2017mean}, UA-MT~\cite{yu2019uncertainty}, DTC~\cite{luo2021semi}, CT~\cite{luo2022semi}, MC-Net~\cite{wu2021semi}, MC-Net+~\cite{wu2022mutual}, URPC~\cite{luo2022URPC}, MCF~\cite{wang2023mcf}, SCP-Net~\cite{zhang2023self}, CauSSL~\cite{miao2023caussl}, BS-Net~\cite{he2023bilateral}, CDMA~\cite{zhong2023semi}, UPCoL~\cite{lu2023upcol}, BCP~\cite{bai2023bidirectional}, and CAML \cite{gao2023correlation}. 
To comprehensively evaluate the effectiveness of our model, we conduct extensive experiments on seven diverse segmentation tasks. These tasks consist of both 2D and 3D medical imaging data acquired from various modalities, such as  MRI, CT, ultrasound, and colonoscopy. 

For the evaluation of the 2D dataset, we use multiple metrics including the Dice coefficient, Hausdorff distance (95HD), Intersection over Union (IoU), and E-Measure ($E_{M}$)~\cite{fan2018enhanced}. For the 3D dataset, we replace $E_{M}$ with Average Surface Distance (ASD).
Besides, to ensure the fairness of our experiments, we maintain identical labeled and unlabeled training datasets. Moreover, we apply the same data augmentation techniques across all datasets.

\begin{figure*}[!t]
\centering
\includegraphics[width=0.96\linewidth]{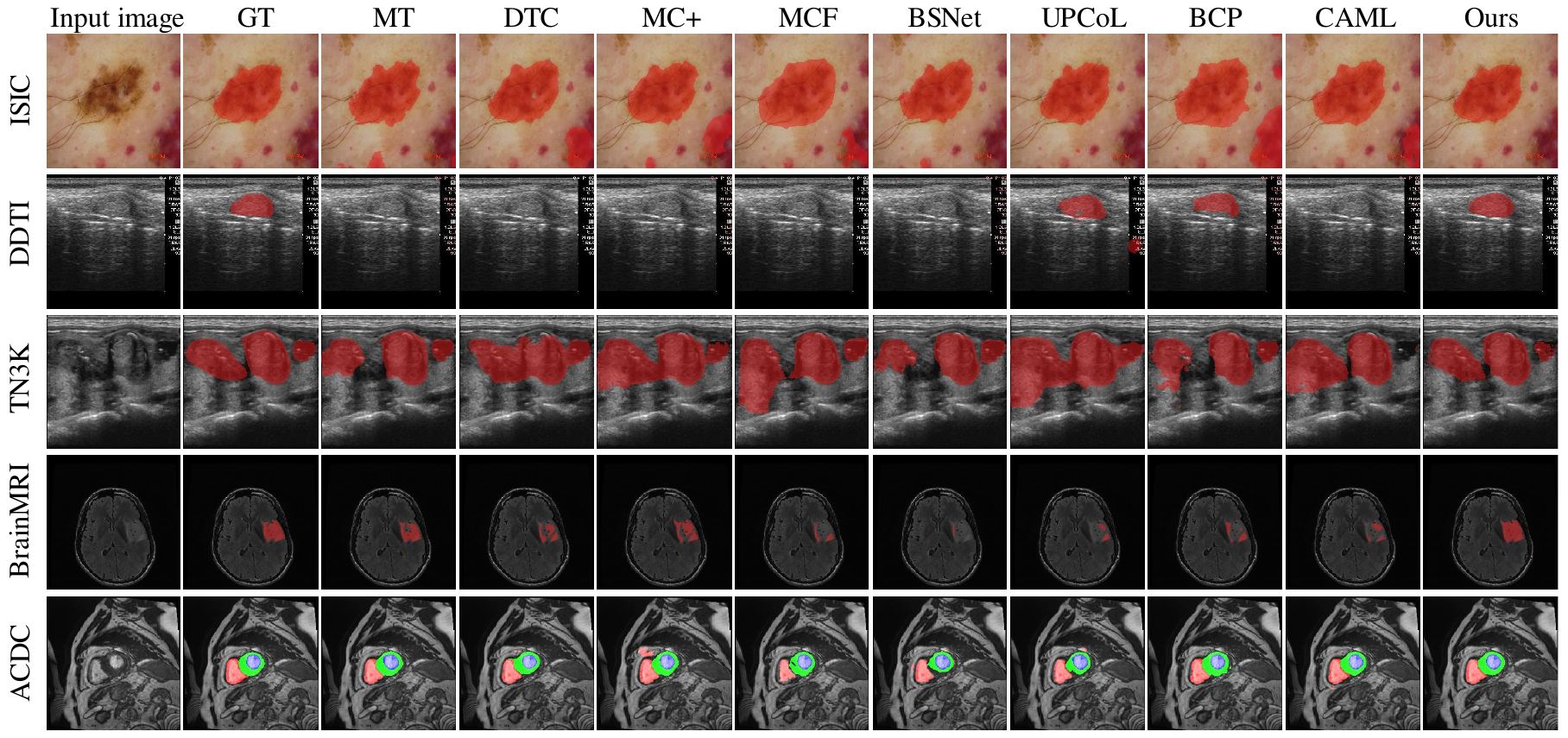}\vspace{-0.25cm}
\caption{{Experimental evaluation of various methods on four additional 2D medical datasets with $10\%$ labeled data}. 
}
\label{fig:other_2D}\vspace{-0.05cm}
\end{figure*}

\subsubsection{Results on Polyp Segmentation} 
\textbf{Learning Ability}. 
We conduct two experiments to validate our model's learning ability on two seen datasets, \ie, CVC-ClinicDB and Kvasir. As shown in Table \ref{tab:tab1}, when trained with only $30\%$ labeled data, our method outperforms BCP on the CVC-ClinicDB dataset, with the dice coefficient improvement from $84.21\%$ to $87.10\%$, and the performance of IoU improvement from $76.64\%$ to $81.09\%$. Compared to CAML, the 95HD coefficient decreases from $3.67$ to $3.36$. When the labeled data decreased to $10\%$, our model achieves a comprehensive superiority over other methods. 
The advantage exhibited by our method stems from its effective incorporation of uncertainty. Through the utilization of uncertainty maps, our approach is capable of making a reliable fusion of the segmentation maps from the two subnets, leading to superior predictions specifically at low-contrast tissue boundaries. 
Moreover, as shown in Fig.~\ref{fig:tumor_result}, the results clearly demonstrate that our model generates superior segmentation maps compared to other semi-supervised methods. It is worth noting that our model is particularly effective in accurately delineating the connection areas of small and multiple targets, a challenging aspect where alternative algorithms often suffer from missed or false detections.

\textbf{Generalization}. To validate the model's generalizability, we conduct comparison experiments on three unseen datasets, namely CVC-300, CVC-ColonDB, and ETIS. As shown in Table \ref{tab:tab1}, our model still achieves the best segmentation performance compared to other methods. Specifically, results from the $10\%$ labeled ETIS dataset indicate that our method achieves an enhancement from $46.1\%$ to $51.7\%$ in terms of Dice coefficients compared to BCP. When the labeled data is increased to $30\%$, although the gap is narrowed, our method still demonstrates excellent generalization performance even with a small amount of labeled data.

\subsubsection{Results on Other 2D Datasets}

To further validate the effectiveness of the proposed model, we employ more comparison experiments on other four 2D medical image tasks, comprising a total of five test sets. We present a comparison of results on a skin dataset in Fig.~\ref{fig:other_2D}, demonstrating the ability of our method to accurately segment inconspicuous regions. When evaluated with only $10\%$ and $30\%$ labeled data, compared to the second-best model, we observe a significant decrease in HD95 values from $4.96$ and $4.84$ to $4.80$ and $4.52$, respectively.
These findings indicate that our method excels in the segmentation of contour edges, resulting in a precise delineation of object boundaries. Overall, the comparison results on the skin dataset showcase the effectiveness of our method in accurately capturing the intricate details of inconspicuous regions and improving the segmentation precision of contour edges.

Fig.~\ref{fig:other_2D} display the results on two ultrasound datasets (DDTI and TN3K), where ultrasound data exhibits lower contrast compared to other medical images. As there is no specific training set available for the DDTI dataset, the model relies on inference results obtained from training on TN3K and TG3K datasets. It is important to note that while the performance of all semi-supervised models falls short of complete satisfaction, this can be attributed to the limited availability of annotated data from other devices for inferring models on previously unseen ultrasound images. Nevertheless, the results obtained on the DDTI dataset demonstrate that our method outperforms other models in cross-device image scenarios. Additionally, the proposed method still outperforms other semi-supervised methods on the TN3K test set, especially in multi-objective scenarios, where it is capable of handling complex segmentation tasks involving multiple regions of interest.
These results highlight the effectiveness and robustness of the proposed model in handling various imaging scenarios. 
Moreover, additional experiments are conducted on the brain gliomas and ACDC datasets, with specific data presented in Table \ref{tab:tab2}. 
Furthermore, Figs.~\ref{fig:other_2D} demonstrate the results on the two datasets with $30\%$ labeled data. It can be observed that our method still performs better than other compared methods, meanwhile highlighting the precise capability in segmenting small target regions.

\subsubsection{Comparison on LA and NIH Pancreas Datasets}

\begin{figure}%[!t]
\centering
\includegraphics[width=0.99\linewidth]{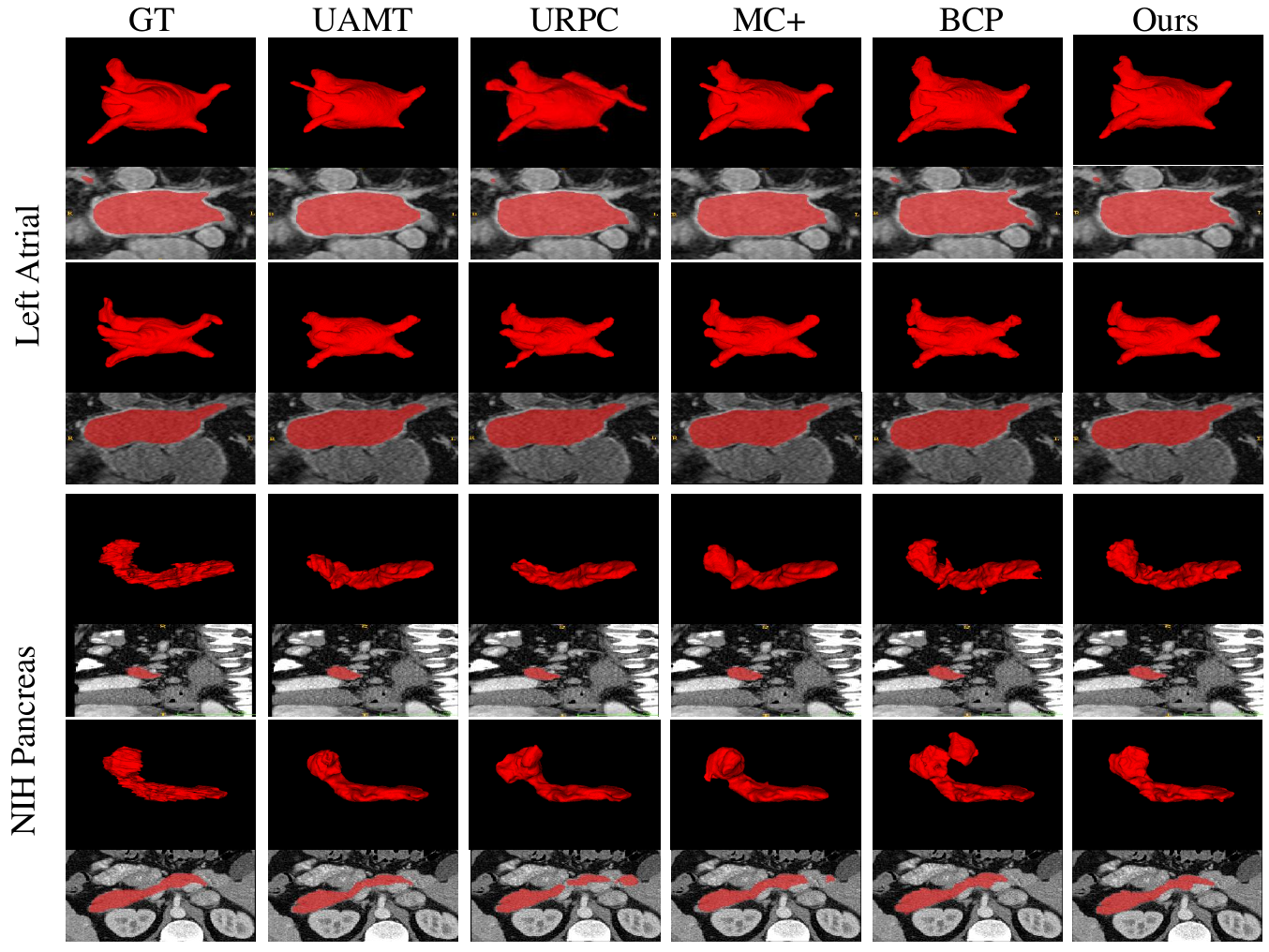}\vspace{-0.15cm}
\caption{Performance evaluation of our method on the two 3D datasets. 
}\vspace{-0.15cm}
\label{fig:3D_result}
\end{figure}

To further demonstrate the generalization of the proposed semi-supervised method, we conduct additional experiments on 3D data. We evaluate the capabilities of our method on volumetric medical images using the Left Atrial (LA) Dataset. 
To facilitate comparison with previous methods, we carry out experiments with $5\%$ and $10\%$ labeled data. 
As shown in Table \ref{tab:3D_Res}, our method outperforms BCP in terms of various evaluation metrics. Specifically, the Dice coefficient improved from $88.02\%$ to $90.36\%$, while the Jaccard coefficient increases from $78.72\%$ to $82.48\%$. Additionally, the HD95 coefficient decreases from $7.90$ to $6.35$, and the ASD value decreases from $2.15$ to $1.76$. Compared to these semi-supervised methods, our model consistently achieves the most favorable results across all evaluation metrics, further validating its superiority in this context. Fig.~\ref{fig:3D_result} shows the visual comparison results on the Left Ventricle segmentation. It can be observed that our method excels in segmenting small synapses, demonstrating a superior level of detail awareness in contrast to other models. As a result, our method exhibits fewer segmentation errors, indicating its enhanced accuracy and precision. Moreover, corresponding 2D slices of the Left Atrial dataset are included in the presentation, further highlighting the strengths of our method. These slices accentuate the superior performance of our approach in accurately segmenting elongated targets compared to other semi-supervised models. 
Our method significantly outperforms other semi-supervised methods in all metric evaluations, with no outliers in terms of Dice coefficient and Miou, indicating stable predictive performance on this dataset. 

\begin{figure*}[!t]
\centering
\includegraphics[width=1\linewidth]{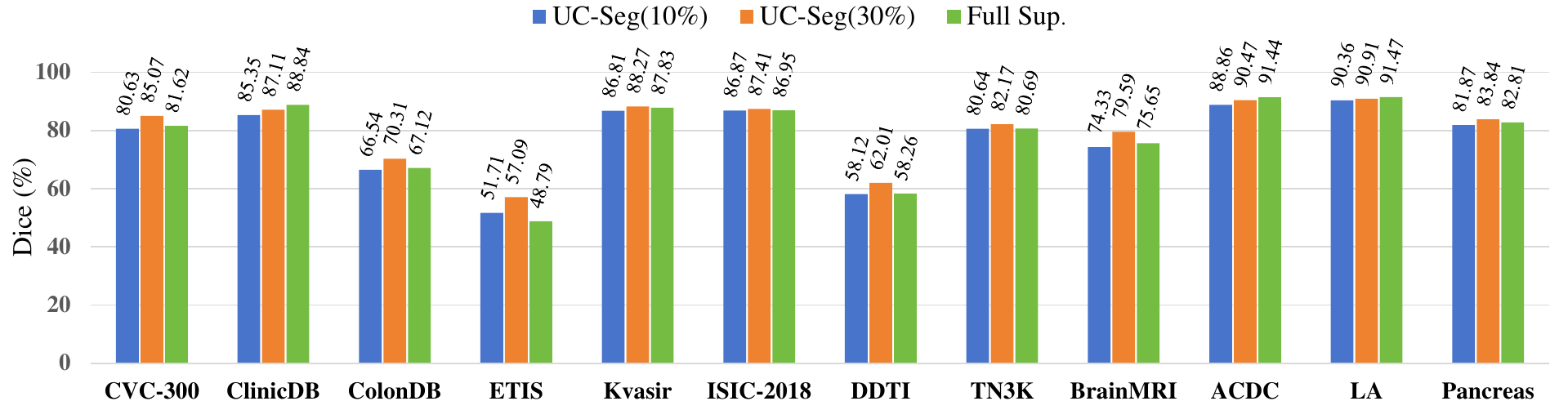}\vspace{-0.25cm}
\caption{Comparative results between our UC-Seg with $10\%$ or $30\%$ labeled data and fully-supervised methods, where U-Net is used for 2D datasets and V-Net is used for 3D datasets, with $100\%$ labeled data.}
\label{fig:Abi_Unet_Vnet}\vspace{-0.15cm}
\end{figure*}

Additionally, we conduct comparison experiments on the NIH Pancreas dataset. From the results in Fig.~\ref{fig:3D_result}, it is evident that other semi-supervised models are more susceptible to producing under-segmentation and over-segmentation results. This implies that these models struggle to accurately delineate boundaries between different regions within the segmentation task. 
In Table \ref{tab:3D_Res}, it can be seen that our method surpasses the BCP model by achieving a notable improvement of $3.38\%$ and $1.53\%$ in mIoU under $10\%$ and $20\%$ labeled data, respectively. 
These results emphasize the superiority of our method in achieving more accurate and reliable segmentations compared to other methods.

\subsubsection{Comparison with Fully-supervised Methods}

In addition, we conduct comparative experiments with the fully-supervised algorithms U-Net (for 2D datasets) and V-Net (for 3D datasets) across seven segmentation tasks. As shown in Fig.~\ref{fig:Abi_Unet_Vnet}, our method achieves comparable performance with the supervised U-Net and V-Net models using only $10\%$ of labeled data.

\subsection{Model Study}

\subsubsection{Ablation Studies}

\begin{table*}
 \renewcommand{\arraystretch}{1.1}
 \setlength\tabcolsep{3.2pt}
    \centering
    \normalsize
    \small
    \caption{Ablation study on the three key modules in the proposed UC-Seg model.
    }\vspace{-0.15cm}
    \begin{tabular}{c|cccc|cc|cc|cc|cc|cc|cc|cc}
    \hline
        \multirow{2}*{} & \multicolumn{4}{c|}{{\textbf{Settings}}}  & \multicolumn{2}{c|}{Kvasir} & \multicolumn{2}{c|}{ISIC}  & \multicolumn{2}{c|}{BrainMRI} & \multicolumn{2}{c|}{TN3K} & \multicolumn{2}{c|}{ACDC}& \multicolumn{2}{c|}{LA}& \multicolumn{2}{c}{Pancreas}\\
        \cline{2-19}
         & Baseline & UPG & IFE & IC  & Dice & 95HD & Dice & 95HD & Dice & 95HD & Dice & 95HD & Dice & 95HD & Dice & 95HD & Dice & 95HD\\
        \hline
        No.1  & \checkmark & $\times$ & $\times$ & $\times$  &  80.57  & 4.58 & 81.86  & 4.84 & 63.03  & 3.57 & 67.35  & 5.36 & 83.32 & 2.99 & 75.23 & 19.08 & 63.10 & 21.74\\
        No.2  & \checkmark & \checkmark &$\times$ & $\times$   &  84.56  & 4.17 & 84.45  & 4.68 & 66.18  & 3.56 & 75.34  & 5.01 & 86.81 & 3.35 & 87.57 & 10.02 & 72.76 & 20.24\\
        {No.3}  & \checkmark & $\times$ &\checkmark & $\times$   & 84.97  & 4.14 & 84.98 & 4.62 & 65.56 & 3.51 & 76.90 & 4.86 & 87.28 & 3.10 & 86.75 & 12.50 & 74.04 & 13.07 \\
        {No.4}  & \checkmark & $\times$ &$\times$ &  \checkmark  & 84.53  & 4.11 & 85.11 & 4.68 & 67.15 & 3.45 & 76.49 & 4.86 & 87.53 & 3.15 & 84.72 & 13.17 & 71.25 & 16.76\\
        No.5  & \checkmark & \checkmark & \checkmark & $\times$   &  85.54  & 4.22 & 85.72  & 4.64 & 71.16  & 3.36 & 78.64  & 4.78 & 87.91 & 3.46 & 88.52 & 9.47 & 80.43 & 6.76\\
        {No.6}  & \checkmark & \checkmark &$\times$ &  \checkmark  & 86.17  & 4.14 & 86.18 & 4.67 & 67.88 & 3.43 & 78.86 & 4.68 & 88.16 & 2.88 & 89.28 & 8.02 & 77.54 & 7.64\\
        {No.7}  & \checkmark & $\times$ & \checkmark&  \checkmark  & 86.40  & 4.16 & 86.09 & 4.57 & 67.41 & 3.42 & 80.36 & 4.58 & 87.87 & 2.52 & 87.37 & 10.38 & 74.06 & 15.99\\
        No.8  & \checkmark & \checkmark & \checkmark & \checkmark  & \textbf{86.81}  & \textbf{4.10} & \textbf{86.37}  & \textbf{4.62} & \textbf{74.33}  & \textbf{3.23} & \textbf{80.64}  & \textbf{4.53} & \textbf{88.86} & \textbf{1.58} & \textbf{90.36} & \textbf{6.35} & \textbf{81.71} & \textbf{4.99} \\
        \hline
\end{tabular}\label{Ablation tab}\vspace{-0.25cm}
\end{table*}

We conduct experiments to investigate the effectiveness of each key component in the proposed model. The ablation study is conducted on all segmentation tasks with $10\%$ labeled data, and the comparison results are shown in Table~\ref{Ablation tab}. Notably, in the absence of the UPG module, we utilize a method where the two subnets provide mutual supervision. The findings reveal a significant reduction in segmentation accuracy upon removal of the UPG module, particularly in challenging segmentation tasks. For instance, on the LA dataset, the dice coefficient drops significantly from $90.36\%$ to $87.37\%$.
Furthermore, after incorporating the IC and IFE components into the feature layers, performance has been further enhanced. These results indicate that applying a soft consistency constraint at the feature level helps prevent each subnet from developing biases toward the input data, thereby improving segmentation performance. More significantly, the interaction between subnets and within each subnet ensures that the learned features and predictions are more reliable and robust.

\subsubsection{Uncertainty Estimation}

\begin{figure}[!t]
\centering
\includegraphics[width=0.95\linewidth]{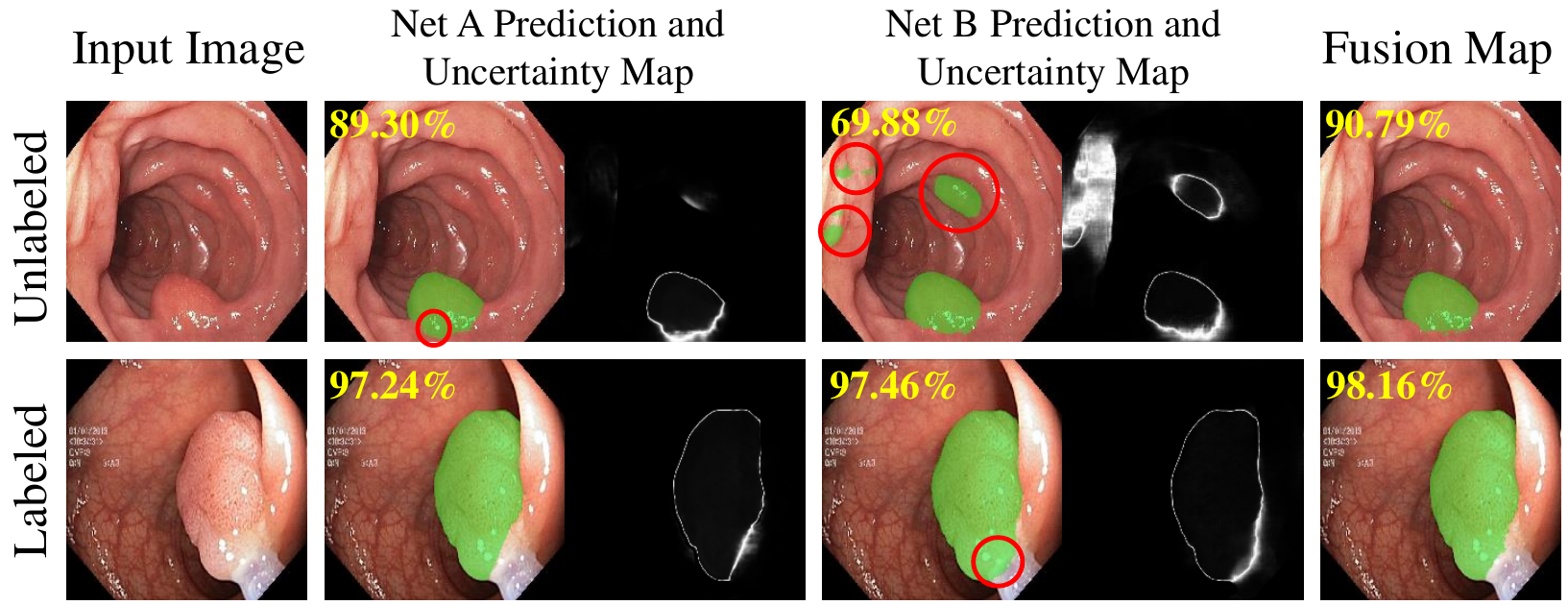}\vspace{-0.15cm}
\caption{Comparison of predictions and uncertainty maps generated by each sub-network for both labeled and unlabeled data, along with the fused predictions. Red circles highlight prediction errors, and yellow numbers indicate the Dice values.
}%\vspace{-0.05cm}
\label{fig:Abi_uncertainty}
\end{figure}

Uncertainty estimation plays a crucial role in extracting the confidence region from the segmentation maps generated by two subnets on unlabeled data. This estimation allows our model to discern between reliable and uncertain regions within the segmentation. When the base network is trained with a small amount of annotated data, it results in semi-confident regions in test images. The degree of confidence for these regions typically falls within the range $0.4$ to $0.7$ (with a maximum value of 1). The proposed method can incorporate the uncertainty maps of individual sub-networks and their position encoding to generate results with high confidence. As shown in Fig.~\ref{fig:Abi_uncertainty}, the white pixels represent highly uncertain/low-confidence regions in the confidence map, while black pixels represent certain/high-confidence regions. 
It can be observed that the sub-network produces a significant amount of ambiguous regions around tumors or tumor-like areas. Our method effectively minimizes the presence of uncertain regions, thereby producing a segmentation mask with predominantly high confidence. We also performed ablation experiments with different fusion strategies, namely ``Avg." and ``Concat.", as shown in Table~\ref{fusion strategies tab}. In the ``Avg." strategy, the outputs of the two subnets are averaged after summing. While in ``Concat.", the outputs are concatenated and passed through a convolutional layer. The results indicate that our proposed UPG strategy with cross-attention outperforms the other two strategies. By effectively integrating uncertainty maps and position encodings, it achieves better alignment and higher confidence in the final segmentation.
Overall, uncertainty estimation provides a mechanism for our model to gain a deeper understanding of the segmentation task by distinguishing between reliable and uncertain regions. This knowledge can be leveraged to make more informed decisions and further boost the segmentation performance.

\begin{figure}[!t]
\centering
\includegraphics[width=0.8\linewidth]{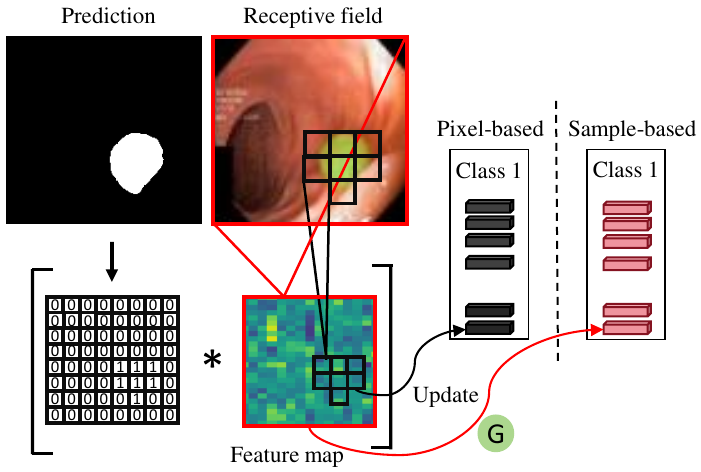}%\vspace{-0.35cm}
\caption{Comparison of pixel-based and sample-based prototype strategies.
}\vspace{-0.25cm}
\label{fig:prototype}
\end{figure}

\begin{table*}[!t]
 \renewcommand{\arraystretch}{1.0}
 \renewcommand{\tabcolsep}{2.3mm}
    \centering
    \caption{Ablation study on different fusion strategies for sub-networks.}\vspace{-0.15cm}
    %\begin{adjustbox}{width=0.5\textwidth}
    \begin{tabular}{c|cc|cc|cc|cc|cc|cc|cc}
    \hline
       \multirow{2}*{Methods} & \multicolumn{2}{c|}{Kvasir} & \multicolumn{2}{c|}{ISIC} & \multicolumn{2}{c|}{BrainMRI} & \multicolumn{2}{c|}{TN3K} & \multicolumn{2}{c|}{ACDC} & \multicolumn{2}{c|}{LA} & \multicolumn{2}{c}{Pancreas}\\
       \cline{2-15}
       & Dice  & 95HD  & Dice & 95HD  & Dice & 95HD  & Dice & 95HD  & Dice & 95HD  &Dice & 95HD &Dice & 95HD \\
       \hline
    Avg.  & 85.18 & 4.18 & 85.61 & 4.62 & 68.33 & 3.41 & 79.41 & 4.58 & 88.27 & 2.90 & 85.29 & 12.55 & 65.01 & 17.87\\
    Concat.  & 84.25 & 4.16 & 85.27 & 4.72 & 66.82 & 3.47 & 74.94 & 5.11 & 87.69 & 4.54 & 84.08 & 11.96 & 65.72 & 17.49\\
    Cross Att.  & \textbf{86.81}  & \textbf{4.10} & \textbf{86.37}  & \textbf{4.62} & \textbf{74.33}  & \textbf{3.23} & \textbf{80.64}  & \textbf{4.53} & \textbf{88.86} & \textbf{1.58} & \textbf{90.36} & \textbf{6.35} & \textbf{81.71} & \textbf{4.99}\\
    \hline
    \end{tabular}
\label{fusion strategies tab}
%\end{adjustbox}
\end{table*}

\begin{table*}%[!t]
 \renewcommand{\arraystretch}{1.0}
 \renewcommand{\tabcolsep}{2.2mm}
    \centering
    \caption{Ablation study on comparing the sample-based prototype with pixel-based prototype.}\vspace{-0.2cm}
     %\begin{adjustbox}{width=0.5\textwidth}
    \begin{tabular}{c|cc|cc|cc|cc|cc|cc|cc}
    \hline
       \multirow{2}*{Settings} & \multicolumn{2}{c|}{Kvasir} & \multicolumn{2}{c|}{ISIC} & \multicolumn{2}{c|}{BrainMRI} & \multicolumn{2}{c|}{TN3K} & \multicolumn{2}{c|}{ACDC} & \multicolumn{2}{c|}{LA} & \multicolumn{2}{c}{Pancreas}\\
       \cline{2-15}
       & Dice  & 95HD  & Dice & 95HD  & Dice & 95HD   & Dice & 95HD  & Dice & 95HD & Dice & 95HD & Dice & 95HD  \\
       \hline
   Pixel-based  & 85.61 &  4.12 & 85.98 & 4.72 & 72.04 & 3.34 & 80.04 & 4.57 & 87.89 & 2.66 & 89.53 & 7.85 & 75.82 & 14.17\\
   Sample-based & \textbf{86.81}  & \textbf{4.10} & \textbf{86.37}  & \textbf{4.62} & \textbf{74.33}  & \textbf{3.23} & \textbf{80.64}  & \textbf{4.53} & \textbf{88.86} & \textbf{1.58} & \textbf{90.36} & \textbf{6.35} & \textbf{81.71} & \textbf{4.99}\\
    \hline
    \end{tabular}
\label{prototype tab}
 %\end{adjustbox}
\end{table*}

\begin{table*}[!t]
 \renewcommand{\arraystretch}{1.0}
 \renewcommand{\tabcolsep}{2.3mm}
    \centering
    \caption{Ablation study on different consistency strategies.}\vspace{-0.15cm}
    %\begin{adjustbox}{width=0.5\textwidth}
    \begin{tabular}{c|cc|cc|cc|cc|cc|cc|cc}
    \hline
       \multirow{2}*{Methods} & \multicolumn{2}{c|}{Kvasir} & \multicolumn{2}{c|}{ISIC} & \multicolumn{2}{c|}{BrainMRI} & \multicolumn{2}{c|}{TN3K} & \multicolumn{2}{c|}{ACDC} & \multicolumn{2}{c|}{LA} & \multicolumn{2}{c}{Pancreas}\\
       \cline{2-15}
       & Dice  & 95HD  & Dice & 95HD  & Dice & 95HD  & Dice & 95HD  & Dice & 95HD  &Dice & 95HD &Dice & 95HD \\
       \hline
   MSE  & 85.92 &  4.15 & 86.19 & 4.76 & 70.49 & 3.32 & 80.40 & \textbf{4.48} & 88.07 & 2.64 & 89.66 & 7.92 & 80.12 & 6.29\\
    CL & \textbf{86.81}  & \textbf{4.10} & \textbf{86.37}  & \textbf{4.62} & \textbf{74.33}  & \textbf{3.23} & \textbf{80.64}  & {4.53} & \textbf{88.86} & \textbf{1.58} & \textbf{90.36} & \textbf{6.35} & \textbf{81.71} & \textbf{4.99}\\
    \hline
    \end{tabular}
\label{consistency tab}
%\end{adjustbox}
\end{table*}

\subsubsection{Effectiveness of Sample-based Prototype}

The acquisition methods for sample-based prototypes and pixel-based prototypes\cite{du2022weakly} are illustrated in Fig.~\ref{fig:prototype}. In this study, we adopt a sample-based prototype approach for feature bank storage. Sample-based prototypes possess a larger receptive field, allowing for a more macro-level exploration of semantic features. Moreover, sample-based prototypes are more suitable for the network structure in this study, as within the IC module, we employ a local patch strategy for consistency regularization. Here, we conduct an ablation experiment with pixel-based prototypes on feature maps. As shown in Table~\ref{prototype tab}, on the 2D unseen ETIS dataset, the dice coefficient increases from $54.24\%$ to $57.09\%$. Similarly, on the 3D LA dataset, the sample-based strategy performs better than the pixel-based method.

\subsubsection{Effectiveness of Contrastive Learning-based Consistency}

We further validate the effectiveness of the adopted contrastive learning consistency strategy. {As shown in Table~\ref{consistency tab}, on the LA dataset with $10\%$ labeled data, the dice coefficient increases from $89.66\%$ to $90.36\%$, and the 95HD decreased from $7.92$ to $6.35$. Compared to directly employing Mean Squared Error (MSE), our contrastive learning consistency not only promotes similarity within the same region but also treats all patches from other positions within a batch as negative samples. This strategy places greater emphasis on the semantic exploration of details, thereby enhancing the overall performance of the model.

\section{Conclusion}
\label{Conclusion}

In this paper, we have proposed a novel semi-supervised network framework, UC-Seg, for medical image segmentation. UC-Seg consists of two distinct sub-networks and an Uncertainty-aware Pseudo-label Generation (UPG) module. In the encoder stages of the two sub-networks, global features for each class are constructed using the Intra-subnet Feature Enhancement (IFE) strategy. Additionally, the Inter-subnet Consistency (IC) strategy is employed to leverage contrastive learning consistency to align the feature maps of the two sub-networks. Finally, to obtain highly reliable outputs, the UPG module integrates the predictions of the two sub-networks by analyzing the uncertainty maps of both networks as guidance. Extensive experimental results on diverse medical segmentation tasks demonstrate the effectiveness of our model over other state-of-the-art semi-supervised medical segmentation methods.

\ifCLASSOPTIONcaptionsoff
  \newpage
\fi

\bibliographystyle{IEEEtran}
\bibliography{arixv}

\end{document}